\documentclass[10pt,twocolumn,letterpaper]{article}

\usepackage[final]{cvpr}
\usepackage{times}
\usepackage{epsfig}
\usepackage{graphicx}
\usepackage{amsmath}
\usepackage{amssymb}

\usepackage{booktabs}
\usepackage{adjustbox}
\usepackage{bbding}
\usepackage{makecell}
\usepackage{multirow}

\usepackage[pagebackref=true,breaklinks=true,letterpaper=true,colorlinks,bookmarks=false]{hyperref}

\begin{document}

%%%%%%%%% TITLE

\title{DocAligner: Annotating Real-world Photographic Document Images by \\ Simply Taking Pictures}

\author{Jiaxin Zhang$^{1}$, Bangdong Chen$^{1}$, Hiuyi Cheng$^{1}$, Fengjun Guo$^{2}$, Kai Ding$^{2}$, Lianwen Jin$^{1}$*\\
$^{1}$South China University of Technology, Guangzhou, China\\
$^{2}$IntSig Information Co., Ltd., Shanghai, China\\
{\tt\small \{eejxzhang, cbb5429\}@gmail.com, eechenghiuyi@mail.scut.edu.cn, eelwjin@scut.edu.cn}\\
{\tt\small \{fengjun\_guo, danny\_ding\}@intsig.net}
}

\maketitle

%%%%%%%%% ABSTRACT
\begin{abstract}
Recently, there has been a growing interest in research concerning document image analysis and recognition in photographic scenarios. However, the lack of labeled datasets for this emerging challenge poses a significant obstacle, as manual annotation can be time-consuming and impractical. To tackle this issue, we present DocAligner, a novel method that streamlines the manual annotation process to a simple step of taking pictures. DocAligner achieves this by establishing dense correspondence between photographic document images and their clean counterparts. It enables the automatic transfer of existing annotations in clean document images to photographic ones and helps to automatically acquire labels that are unavailable through manual labeling. Considering the distinctive characteristics of document images, DocAligner incorporates several innovative features. First, we propose a non-rigid pre-alignment technique based on the document's edges, which effectively eliminates interference caused by significant global shifts and repetitive patterns present in document images. Second, to handle large shifts and ensure high accuracy, we introduce a hierarchical aligning approach that combines global and local correlation layers. Furthermore, considering the importance of fine-grained elements in document images, we present a details recurrent refinement module to enhance the output in a high-resolution space. To train DocAligner, we construct a synthetic dataset and introduce a self-supervised learning approach to enhance its robustness for real-world data. Through extensive experiments, we demonstrate the effectiveness of DocAligner and the acquired dataset. Datasets and codes will be publicly available.
\end{abstract}

\vspace{-3mm}
\section{Introduction}
\label{sec:intro}
In recent years, researchers have made significant strides in document image analysis and recognition. While previous studies predominantly focused on clean document images obtained from digital-born sources or flat-bed scanners~\cite{zhong2019publaynet,li2020docbank,li2020Tablebank,pfitzmann2022doclaynet,wang2021towards}, there is a growing interest among researchers in addressing the challenges posed by more realistic photographic scenarios~\cite{park2019cord,ma2018docunet,long2021parsing,vu2021mc}. However, progress in this field has been hindered by the limited availability of labeled photographic data. This data scarcity can be attributed to several reasons. Firstly, automatic labeling methods~\cite{li2020docbank,zhong2019publaynet,li2020Tablebank,chi2019complicated} designed for clean document images are not suitable for photographic scenarios, necessitating costly and time-consuming manual labeling. Secondly, certain tasks such as illumination correction and geometric rectification are extremely challenging to annotate manually.

\begin{figure}[t]
\includegraphics[scale=0.3]{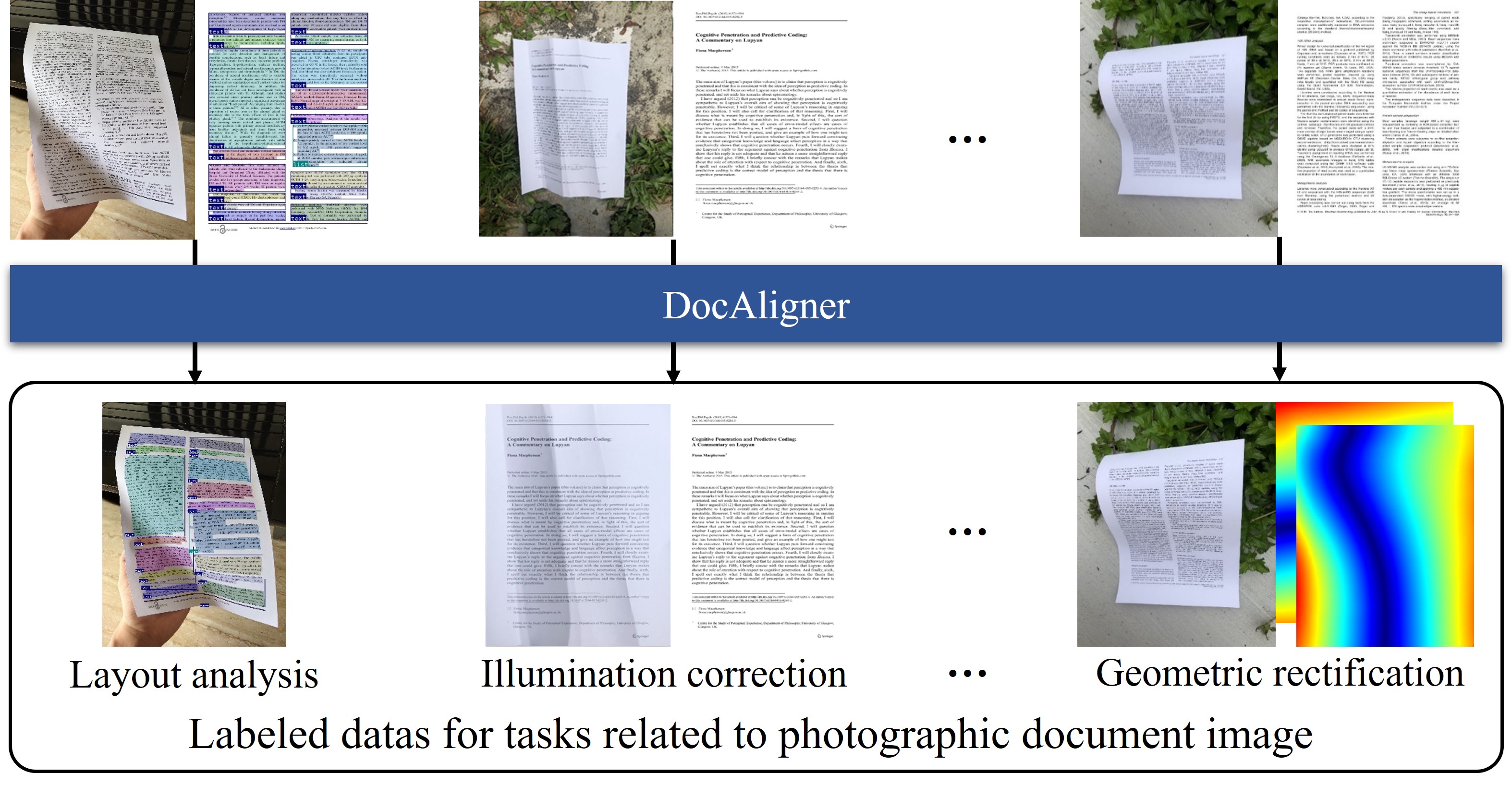}
    \centering
    \caption{Automatic annotation for real-world photographic document images via DocAligner. All you need to do is take pictures.}
    \label{fig:motivation}
    \vspace{-3mm}
    
\end{figure}

To address the aforementioned issues, we propose DocAligner, a novel method that significantly simplifies manual annotation to just taking pictures. DocAligner achieves this by establishing dense correspondence between photographic document images and their clean counterparts, which is a new perspective in the context of document artificial intelligence. As shown in Fig.~\ref{fig:motivation}, for tasks that already have a large amount of labeled clean images (such as layout analysis, table detection, and table recognition), annotations can be transferred to corresponding photographic images. For tasks that cannot rely on existing labeled data and require extensive labeling efforts, we can automatically generate labels following the dense correspondence. In other words, to annotate photographic data, it is only necessary to print a clean document image and take a picture. DocAligner can perform the remaining tasks automatically.

While dense correspondence has been extensively explored in the realm of natural images, utilizing pre-existing models designed for natural images in the context of document analysis faces performance degradation issues resulting from the distribution gap. To address this issue, we make novel designs for DocAligner to achieve dense correspondence exclusively for document images. Pre-alignment becomes necessary for document pairs that exhibit significant global misalignment and contain repetitive patterns. Unlike rigid transformations such as affine and homography commonly used for pre-alignment in natural images, these methods are not suitable for document pairs with non-rigid deformation. Therefore, we propose the use of a thin plate spline (TPS)~\cite{bookstein1989principal} non-rigid transformation, inspired by the advancements in geometric rectification techniques~\cite{zhang2022marior,ma2022learning}. To handle significant shifts and ensure high accuracy, we adopt hierarchical alignment that combines global-local correlation and coarse-to-fine flow prediction. Furthermore, document images possess a more intricate structure, with the details at the character level being crucial. Consequently, obtaining higher-resolution output flows becomes necessary in DocAligner. To address this, we propose a details recurrent refinement module that operates in the detail-rich and high-resolution space. To mitigate memory consumption in high-resolution processing, we conduct refinement recurrently using the memory-efficient ConvGRU~\cite{shi2015convolutional}.

To train DocAligner, we develop a synthetic dataset comprising triplets of photographic document images, clean document images, and flow fields. The clean document images are derived from PDF files. We warp these clean images with randomly-generated flows and then deteriorate them using collected shading maps to synthesize photographic images. Additionally, to further improve DocAligner's performance on real data, we propose a self-supervised learning approach. Experimental results demonstrate the superiority of DocAligner compared to existing methods. Furthermore, we assess the effectiveness of the acquired dataset in multiple tasks related to photographic document images, including layout analysis, illumination correction, and geometric rectification.

In summary, our contributions are as follows:
\begin{itemize}
\item For the first time, we explore the dense correspondence task in the context of document artificial intelligence, by which we ease the data dilemma countered by tasks related to photographic document images.
\item We propose DocAligner for document image dense correspondence, in which we design non-rigid pre-alignment, hierarchical alignment with global and local correlation, and details recurrent refinement. We also develop a synthetic dataset and a self-supervised learning approach that is easy to implement and helps to improve generalization. 
\item DocAligner achieves superior performance compared to existing methods. Additionally, we validate the effectiveness of the dataset we acquired for related tasks.
\end{itemize}

%-------------------------------------------------------------------------
\section{Related works}
\subsection{Dense correspondence}
Dense correspondence of paired images has been extensively studied for natural images in recent years~\cite{ilg2017flownet,sun2018pwc,teed2020raft,rocco2017convolutional,truong2020glu,melekhov2019dgc,han2017scnet,kim2018recurrent}. Given an image pair $(I_s,I_t)$ with a size of $H\times W$, dense correspondence aims to predict a flow field $f\in \mathbb{R}^{H \times W \times 2}$, which relates the source $I_s$ to the target $I_t$. According to differences within the paired images, dense correspondence can be categorized into optical flow~\cite{ilg2017flownet,sun2018pwc,teed2020raft}, geometric correspondence~\cite{rocco2017convolutional,truong2020glu,melekhov2019dgc} and semantic correspondence~\cite{rocco2017convolutional,truong2020glu,han2017scnet,kim2018recurrent}. Document image pairs with large displacements and significant appearance transformations are more relevant to geometric correspondence, where pairs usually exhibit different views of the same scene or are captured by different cameras on different occasions.

Melekhov et al.~\cite{melekhov2019dgc} proposed DGC-Net, a neural network with a global correlation layer that can handle large displacement. However, due to the large memory footprint of this layer, the input image resolution for DGC-Net is constrained to $240\times240$. Such a coarse resolution is insufficient for representing a document with fine-grained content. Glu-Net~\cite{truong2020glu} takes a more elegant approach by performing global correlation in coarse resolution and local correlation in fine resolution, resulting in better performance on high-resolution input. Truong et al. proposed GOCor~\cite{truong2020gocor}, a new optimizable correlation layer, to enhance the correlation robustness of similar and low-textured regions. Some self-supervised methods are also introduced to make models more robust on real-world data. RANSAC-Flow~\cite{shen2020ransac} adopts a two-stage framework where coarse alignment based on RANSAC~\cite{fischler1981random} is followed by a fine alignment based on a deep model trained with self-supervision. However, it is sophisticated and hard to implement due to its multi-task optimization for cycle consistency, matchability, and reconstruction. DMP ~\cite{hong2021deep} optimizes the untrained matching networks on a single pair of images. But it is less practical to focus solely on an input pair. Although achieving promising results on natural images, the above-mentioned methods remain sub-optimal in the context of document images.

\subsection{Document analysis and recognition in photographic scenarios}
\textbf{Document layout analysis (DLA)}.
DLA aims to identify the regions of interest in an unstructured document and determine the role of each region. Previous studies have primarily focused on digital-born document images that are relatively easy to label by parsing PDFs and analyzing the corresponding source codes~\cite{zhang2021vsr,bi2022srrv,yang2017learning,kaplan2021combining} such as LaTeX and XML. However, more and more photographic document images are emerging, which the existing automatic labeling methods cannot cope with. Consequently, while millions of labeled clean document images are available, there are limited datasets for photographic scenarios. Although it is possible to annotate manually (by annotating bounding boxes and classes like title, author, list, abstract, paragraph, table, figure, etc.), it is expensive, especially considering geometric deformation and the large number of objects in photographic document images.

\textbf{Illumination correction}. 
Illumination correction seeks to eliminate degradation caused by uncontrolled illumination, enhancing readability and facilitating following optical character recognition (OCR) engines~\cite{feng2021doctr,li2019document,das2020intrinsic}. Nevertheless, obtaining labels for this task, i.e., the illumination-corrected image, is challenging due to the dense annotation density. An alternative approach to obtain labeled data is to capture the document under different illuminations while keeping its relative position fixed with the camera. However, source documents are not always available, and the variety and scale of datasets obtained in this way are minimal. Thus, most recent learning-based illumination correction methods~\cite{feng2021doctr,li2019document,das2020intrinsic,das2019dewarpnet} can only be trained on synthetic data, whose realism and diversity remain unsatisfactory. Further discussions are included in Section~\ref{illumination}.

\textbf{Geometric rectification}. This task aims to flatten document images that suffer from curves, folds, crumples, etc. A dewarping map is required to sample from distorted input to obtain the rectified result. This dewarping map indicates the correspondence between pixels in the desired rectified results and the distorted inputs. However, such a dewarping map is an extremely dense annotation that can be almost impossible to obtain manually~\cite{xie2021document}. Consequently, many learning-based methods have to resort to synthetic data. Failure to obtain the dewarping map annotation means that real-world data can only be used for weak supervision~\cite{xue2022fourier,ma2022learning}.

\section{Methodology}
\begin{figure*}[t]
\includegraphics[width=5.3in]{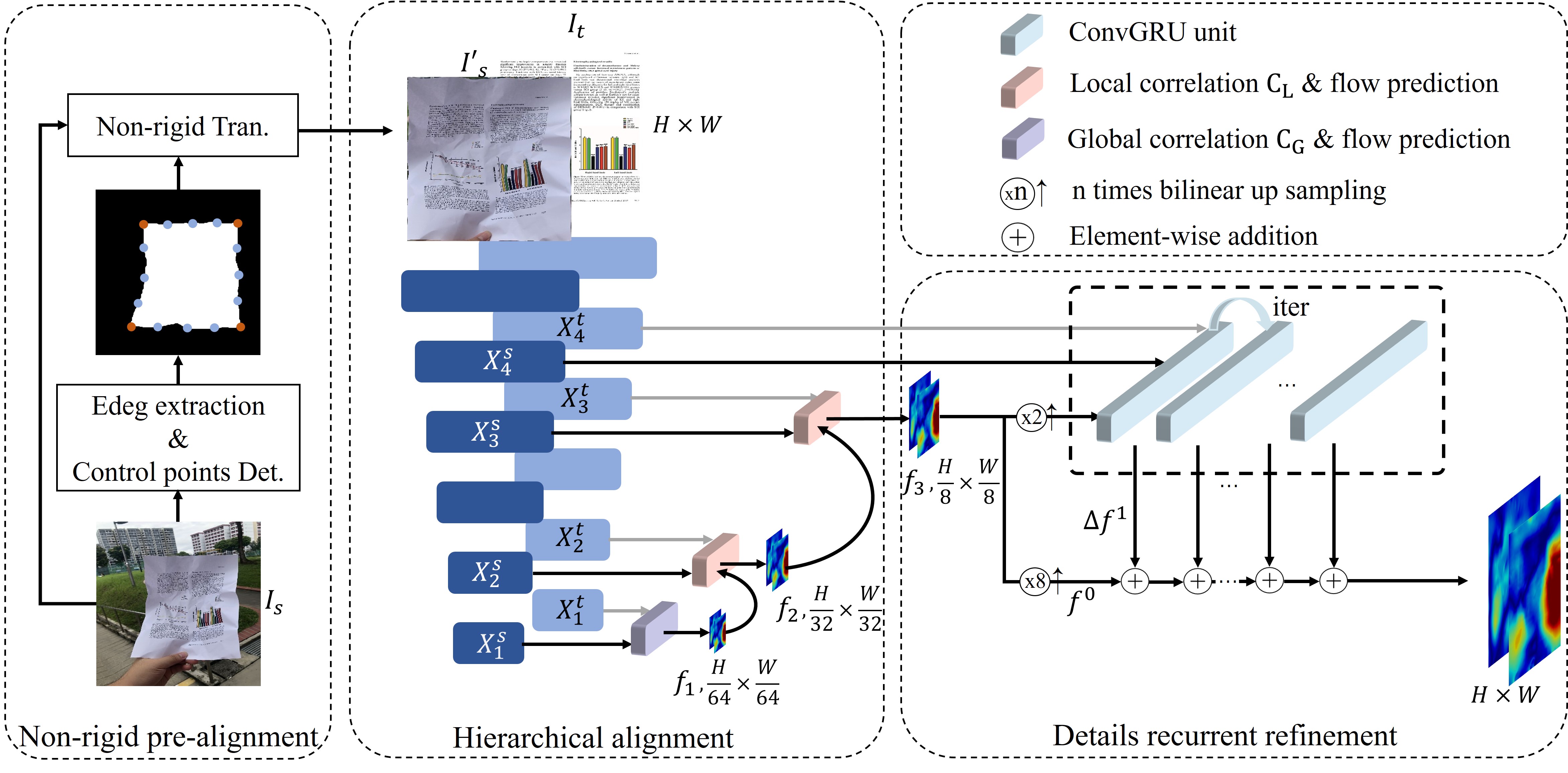}
    \centering
    \caption{Overall architecture of DocAligner. Considering photographic $I_s$ and clean $I_t$, DocAligner aims to densely correlate each pixel in $I_t$ with $I_s$.}
    \label{fig:architecture}
\end{figure*}
As shown in Fig.~\ref{fig:architecture}, the proposed DocAligner seeks to correlate the photographic document image $I_s$ with its clean counterpart $I_t$. To achieve this, a pre-alignment module is first utilized to obtain $I'_s$, which is accomplished using an edge-based non-rigid transformation. Pre-aligned pairs $(I'_s, I_t)$ are then fed into a shared feature extraction backbone to extract multi-scale features, which are then used to predict flows hierarchically. Finally, a refinement module is designed to refine flow details recurrently in high-resolution space and output a flow with the same size as the input pairs.

\subsection{Non-rigid pre-alignement}
Photographic document images often suffer from significant global misalignment caused by the varying camera angles and paper deformations. Such misalignment combined with repetitive patterns in a document impede precise correlations. Pre-aligning the images before carrying out fine-grained correlation can resolve this issue. 
Traditional affine and homography rigid transformations for natural imagess~\cite{rocco2017convolutional,shen2020ransac,jeon2018parn} are not suitable for document pairs that are with both rigid and non-rigid deformations. In this paper, we focus on the advancements in document image rectification which utilize the document edge information~\cite{zhang2022marior}. Specifically, as illustrated in Fig.~\ref{fig:architecture}, we first extract the edge of the document through semantic segmentation. We then detect the four corners and equidistant points on four edges based on edge information. We map these detected points to their pre-defined reference counterparts in a quadrilateral. Using these paired points, we apply the TPS non-rigid transformation to obtain pre-aligned $I'_{s}$.

\subsection{Hierarchical alignment}
As shown in Fig.~\ref{fig:architecture}, we feed the pre-aligned pairs $(I'_{s},I_{t})$ into a shared feature extraction backbone which procude multi-scale features $X^{s}=\left\{X_{1}^{s}, X_{2}^{s}, X_{3}^{s}, X_{4}^{s}\right\}$ and $X^{t}=\left\{X_{1}^{t}, X_{2}^{t}, X_{3}^{t}, X_{4}^{t}\right\}$ for $I'_{s}$ and $I_{t}$, respectively. In the hierarchical alignment module, we predict and fine-tune the flow from low to high resolution, i.e., from level $l=1$ to $l=3$. The flow $f_l$ between a pair of features $(X_{l}^{s}, X_{l}^{t})$ at level $l$ is calculated by 
\begin{equation}\footnotesize
f_l=\mathbf{up}\left(f_{l-1}\right)+\mathbf{decoder}_l\left(C_l, \mathbf{up}\left(f_{l-1}\right)\right),
\end{equation}
where $\mathbf{up}()$ is a bilinear up-sampling function and $\mathbf{decoder}_l()$ is a lightweight  fully convolution neural network. The detailed architecture of $\mathbf{decoder}_l()$ can be found in the supplementary materials. Furthermore, $C_l$ refers to the correlation map obtained through global or local correlation layer $\mathbf{C_{G/L}}$:
\begin{equation}\footnotesize
C_l=\mathbf{C_{G/L}}\left(\tilde{X}_l^s, X_l^t\right).
\end{equation}
Here $\tilde{X}_l^s$ is obtained by warping $X_l^s$ toward $X_l^t$:
\begin{equation}\footnotesize
\tilde{X}_l^s(x)=X_l^s\left(\mathbf{x}+\mathbf{up}\left(f_{l-1}\right)(\mathbf{x})\right),
\end{equation}
where $\mathbf{x}$ denotes the image coordinate. Additionally, the initial flow $f_0$ is a zero-filled map.

The correlation layer, also known as cost volume, is essential in current state-of-the-art dense correspondence methods, as it represents the similarities between spatial elements in both reference and query features. The global correlation layer calculates the scalar product between each feature vector in the reference features $X^{r} \in \mathbb{R}^{H^r \times W^r \times D}$ and all the vectors in the query features $X^{q} \in \mathbb{R}^{H^q \times W^q \times D}$, as following:
\begin{equation}\small
\mathbf{C}_{\mathrm{G}}\left(X^{r}, X^{q}\right)_{i j}=\left(x_{i}^{r}\right)^{\mathrm{T}} x_{j}^{q},
\end{equation}
where $x_{i}^{r} \in \mathbb{R}^{D}$ and $x_{j}^{q}\in \mathbb{R}^{D}$ are the $i$-th and $j$-th vector in $X^{r}$ and $X^{q}$, respectively. This layer, denoted as $\mathbf{C}_{\mathrm{G}} \in \mathbb{R}^{H^rW^r \times H^qW^q}$, represents similarities between all locations in the reference and query features and can handle large displacements. However, its computational complexity and memory consumption increase quadratically with feature size, rendering it only suitable for low-resolution features. In contrast, the local correlation layer computes the scalar product between vectors within a constrained distance:
\begin{equation}\footnotesize
\mathbf{C}_{\mathrm{L}}\left(X^{r}, X^{q}\right)_{i d}=\left(x_{i}^{r}\right)^{\mathrm{T}} x_{i+d}^{q},\qquad \|d\| \leq R
\end{equation}
where $R$ is the pre-defined constant ratio. $\mathbf{C}_{\mathrm{L}} \in \mathbb{R}^{H^rW^r \times (2R+1)}$ is more computationally efficient but unsuitable for correlating feature pairs with large displacements. In this paper, we apply the global correlation layer to the lowest resolution features (i.e., $l=1$) and the local correlation layer with $R=9$ to the remaining levels, enabling hierarchical alignment that can eliminate large global displacements in low resolution and focus precise correlation in high resolution.

\subsection{Details recurrent refinement}

\begin{figure}[t]
    \includegraphics[width=2.7in]{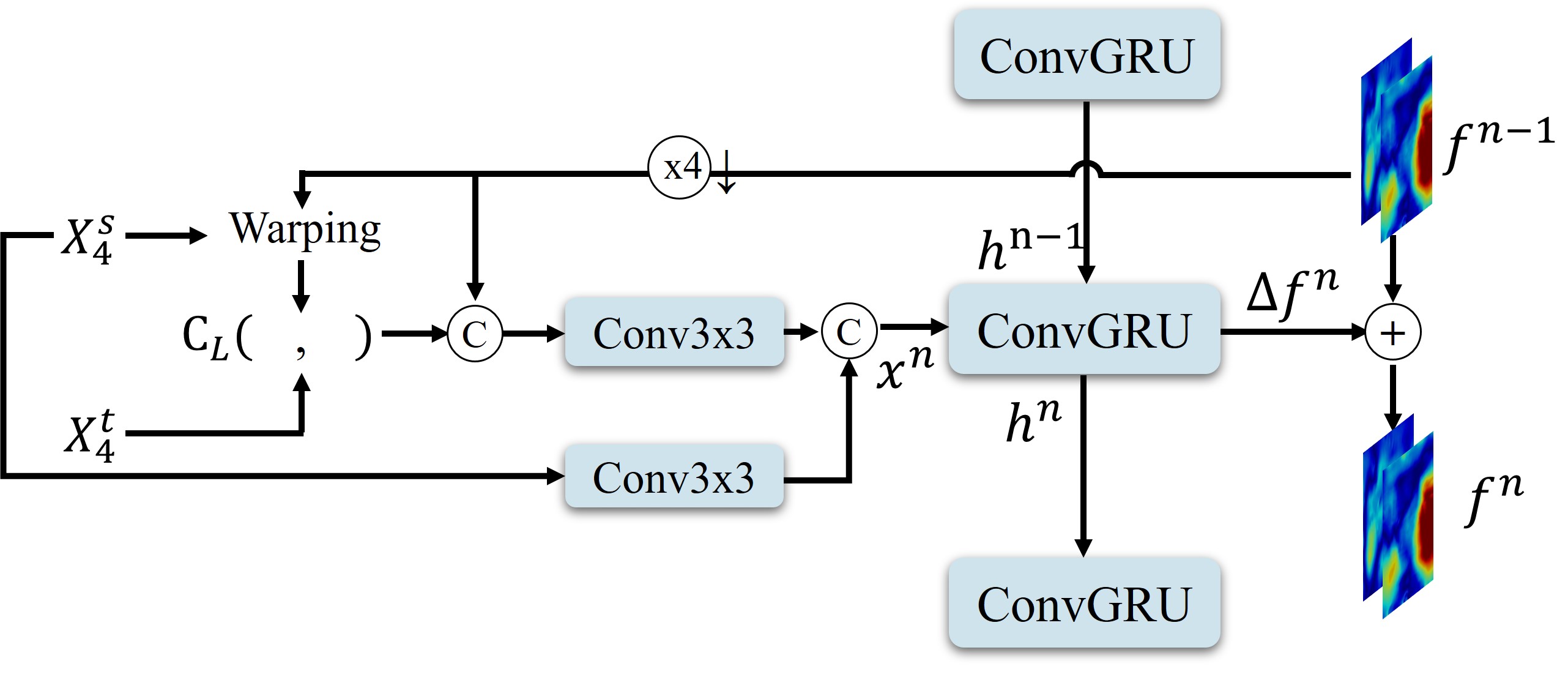}
    \centering
    \caption{The details recurrent refinement module.}
    \label{fig:gru}
\end{figure}

The final output flow field of natural images is usual 1/4 the size of the input image. A flow field with a larger size will not bring much more improvement~\cite{truong2020glu,dosovitskiy2015flownet,sun2018pwc,cho2022cats++}. For some natural scenarios, low-resolution input is enough to obtain good performance~\cite{melekhov2019dgc,shen2020ransac,jiang2021cotr}. Nevertheless, this is not the case for document images because of their fine-grain elements. To address this issue, we propose a details refinement module to obtain the output flow field with the same size as input pairs. As shown in Figs.~\ref{fig:architecture} and~\ref{fig:gru}, this refinement module refines the output in the highest resolution space and employs a recurrent ConvGRU unit~\cite{shi2015convolutional} to reduce memory usage. At each time step $n$, the input of the GRU unit is $x^n$, which is obtained by
\begin{equation}\footnotesize
\centering
x^{n}=[\mathbf{Conv_{3 \times 3}}\left(X_4^s\right) ,motion],
\end{equation}
\begin{equation}\footnotesize
\centering
motion =\mathbf{Conv_{3 \times 3}}\left(\left[\mathbf{down}(f^{n-1}), \mathbf{C_L}\left(\tilde{X}_4^s, X_4^t\right)\right]\right),
\end{equation}
\begin{equation}\footnotesize
\label{eq:8}
\centering
\tilde{X}_4^s(x)=X_4^s\left(\mathbf{x}+\mathbf{down}\left(f^{n-1}\right)(\mathbf{x})\right).
\end{equation}
$\mathbf{Conv_{3 \times 3}}()$ refers to a ${3\times3}$ convolutional layer, $\mathbf{down}()$ is a down-sampling function, and $\tilde{X}_4^s(x)$ is obtained by warping $X_4^s$ toward $X_4^t$.

The GRU unit updates the hidden state from the input $x^n$ and the previous hidden state $h^{n-1}$, similar to a method in \cite{shi2015convolutional}. The updated hidden state $h^{n}$ is used to predict the residual flow $\Delta f^n \downarrow$ using a two-layer convolutional neural network. 

However, the size of $\Delta f^n \downarrow$ ($\frac{H}{4} \times \frac{W}{4}$) is smaller than that of the input image, so we introduce a learnable upsampling method. We predict 16 weight matrices of size $3\times3$ for each pixel in $\Delta f^n \downarrow$, thus obtaining the weight map with size of $\frac{H}{4} \times \frac{W}{4} \times 16 \times 3 \times 3$, which can be reshaped as $\frac{H}{4}  \times \frac{W}{4}  \times 144$. We obtain it by feeding $\Delta f^n \downarrow$ and $h^n$ to another two-layer convolutional neural network (refer to supplementary materials for more details). We then obtain each upsampled pixel using a weighted sum over the $3\times3$ neighborhood in $\Delta f^n \downarrow$ and finally obtain the desired upsampled residual flow $\Delta  f^n$. The updated flow can be obtained by
\begin{equation}\footnotesize
\label{eq:9}
    f^n = f^{n-1} + \Delta  f^{n}.
\end{equation}
Bilinearly up-sampled flow $f_3$ from the hierarchical alignment module is set as the initial flow $f^0$ for Eq.~\ref{eq:8} and ~\ref{eq:9}.  Initial hidden state $h^0$ is transformed from $X_4^s$ through a $1\times1$ convolution layer. We set the iterations for refinement as 7.

% \begin{equation}\small
% \centering
% \begin{aligned}
% &z^n=\sigma\left(\operatorname{Conv}_{3 \times 3}\left(\left[h^{n-1}, x^n\right]\right)\right) \\
% &r^n=\sigma\left(\operatorname{Conv}_{3 \times 3}\left(\left[h^{n-1}, x^n\right]\right)\right) \\
% &\tilde{h}^n=\tanh \left(\operatorname{Conv}_{3 \times 3}\left(\left[r^n \cdot h^{n-1}, x^n\right]\right)\right) \\
% &h^n=\left(1-z^n\right) \cdot h^n+z^n \cdot \tilde{h}^n.
% \end{aligned}
% \end{equation}

\subsection{Self-supervision learning with real data  }
\begin{figure}[h]
    \includegraphics[width=2.8in]{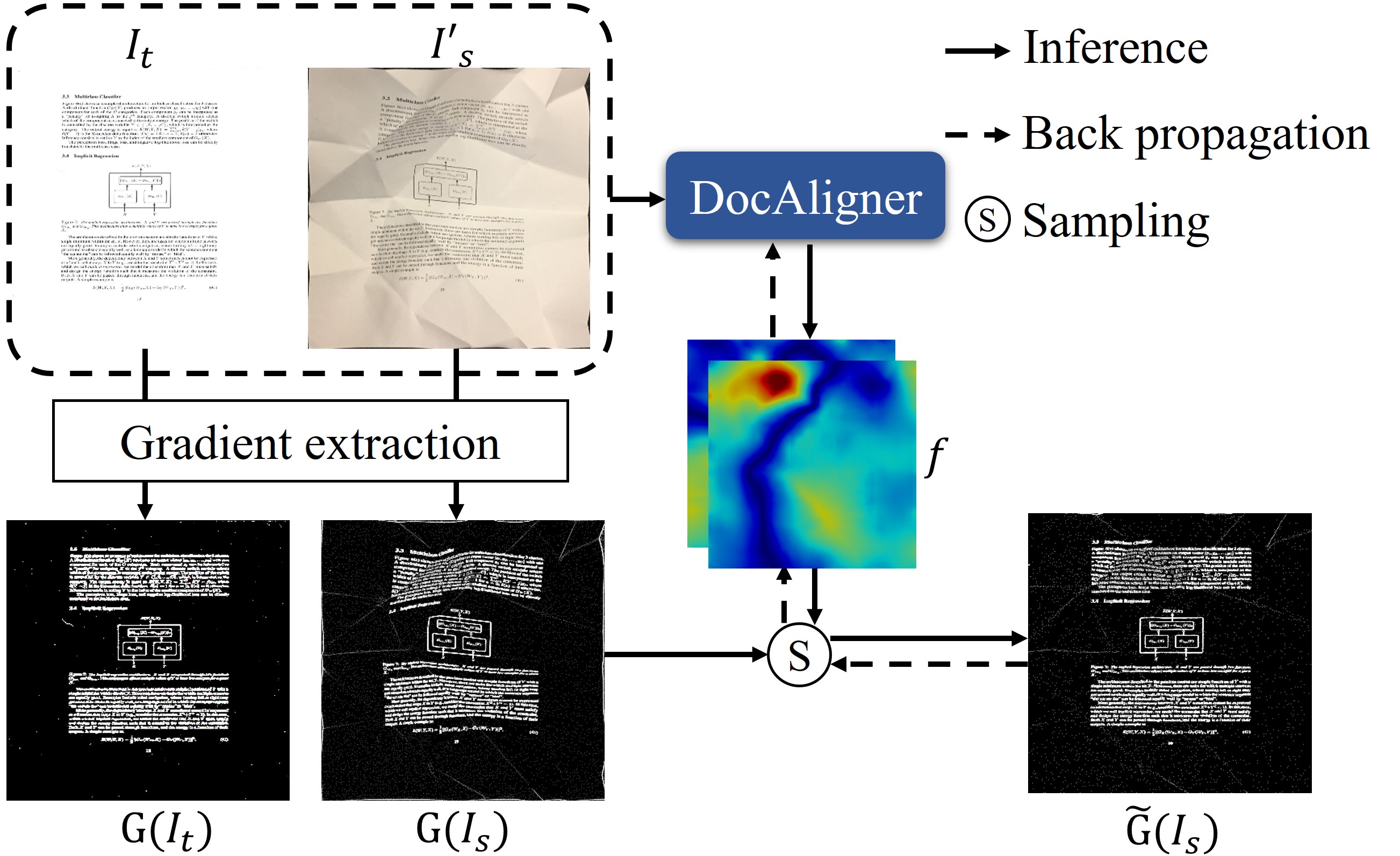}
    \centering
    \caption{Scheme of self-supervision on real data. The dotted lines indicate the back propagation of gradient.}
    \label{fig:selfsupervise}
\end{figure}

We train DocAligner on synthetic data as described in Section~\ref{section:training data}, but we observe a distribution gap between real-world and synthetic data. To tackle this problem , we propose a self-supervised approach to improve DocAligner's robustness, as depicted in Fig.~\ref{fig:selfsupervise}. This approach involves providing DocAligner with paired clean document image $I_{t}$ and pre-aligned photographic image $I'_{s}$ from the real world, and is formulated as 
\begin{equation}\footnotesize
\centering
  \label{self1}
  \hat{\theta}=\underset{\theta}{\operatorname{argmin}}\left(\left\|\tilde{\mathbf{G}}\left(I'_{s}\right)-\mathbf{G}\left(I_{t}\right)\right\|_1\right),
\end{equation}
\begin{equation}\footnotesize
\centering
  \label{self2}
\tilde{\mathbf{G}}\left(I'_{s}\right)=\mathbf{G}\left(I'_{s}\right)\left(\mathbf{x}+f\left(\mathbf{x}\right)\right),
\end{equation}
\begin{equation}\footnotesize
\centering
  \label{self3}
  f=\text {DocAligner }\left(I_{t}, I'_{s} ; \theta\right).
\end{equation}
Here $\theta$ is the parameters of DocAligner, $f$ is the predicted flow field, $\mathbf{x}$ is the image coordinates, and $\mathbf{G}$ is the Sobel operator to extract gradients. It is hard to directly solve Eq.~\ref{self1}, so we approximate it using a gradient descent algorithm. Considering the inefficiency of individual optimization of each sample, similar to prior art~\cite{shen2020ransac}, we optimize our network on the entire test set before testing and then perform inference using the optimized network. It should be noted that our self-supervision process only involves the input data and does not include any ground-truth flow field for supervision.

\section{Experiments}
\subsection{Dataset}
\label{section:training data}
\begin{figure}[h]
    \includegraphics[scale=0.32]{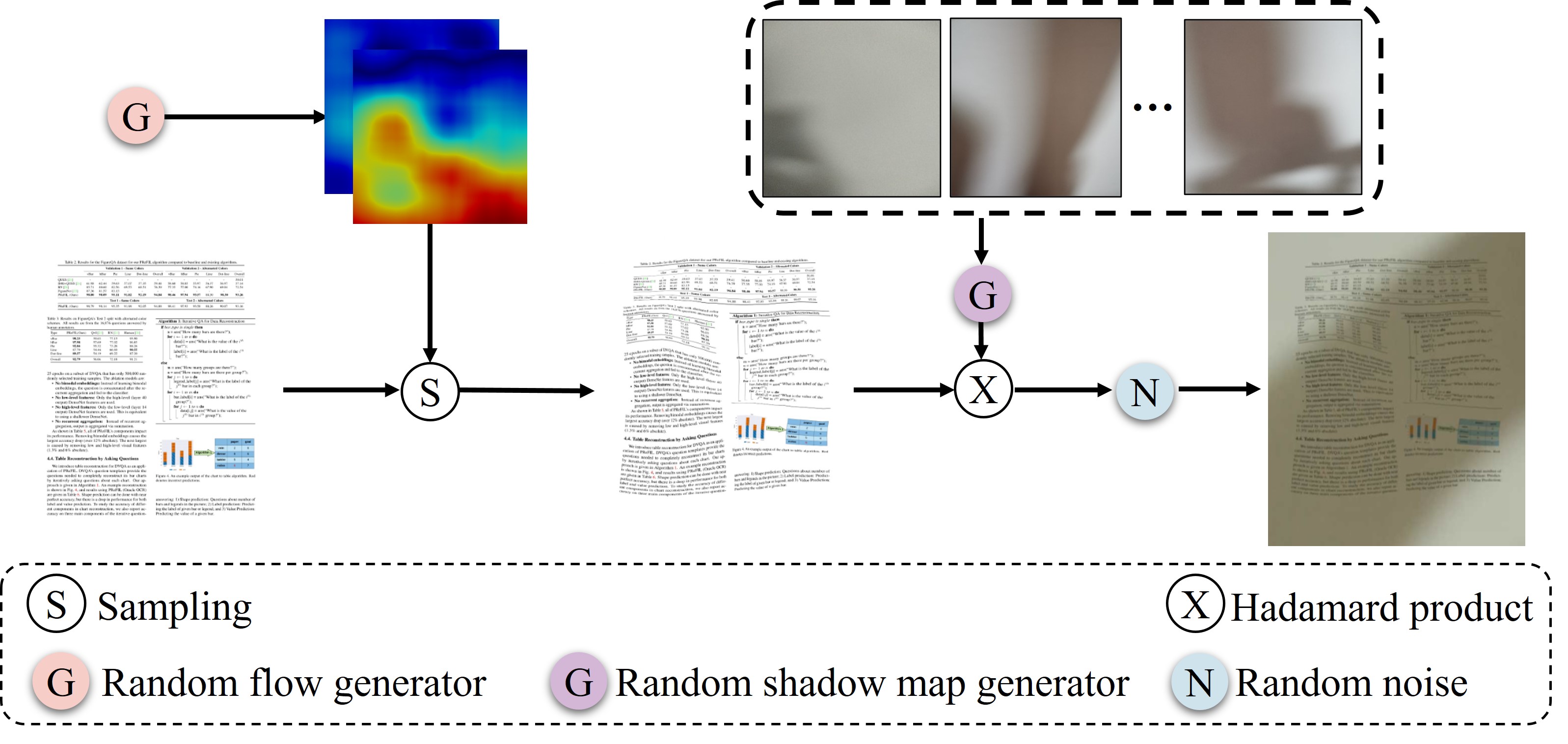}
    \centering
    \caption{The pipeline for synthesizing DocAlign12K.}
    \label{fig:synth}
\end{figure}
Due to the lack of publicly available training data, we develop a synthetic dataset named DocAlign12K, the synthetic pipeline of which is shown in Fig.~\ref{fig:synth}. Firstly, we collect the PDF files from the Internet and then convert them into clean document images. Next, we randomly generate flow fields for each of these images. More detailed procedures and parameter settings are available in the supplementary materials. Using the generated flow fields, we warp clean images to obtain geometrically distorted images. Based on the Lambertian assumption, we consider an image $I$ as a composition of reflectance $R$ and shading $S$, i.e., $I = R \otimes S$, where $\otimes$ denotes the Hadamard product. We collect 500 real shadings by capturing backgrounds without texture under various illumination conditions. We select one of these shadings randomly and perform random cropping, rotation, and color shifts to obtain $S$. Finally, we treat the geometrically distorted image as $R$ to obtain the final image $I$. Additionally, we apply random JPEG compression, Gaussian noise, blur, etc., to better simulate the degradations. We obtain 12K samples (with triplets of clean documents, photographic documents, and flow fields) and split them into training (10K) and testing (2K) sets. 

\subsection{Implementation details}
We implement DocAligner in the PyTorch framework~\cite{pytorch}, and train it on two NVIDIA 2080Ti GPUs with a batch size of 4. The widely-used Adam optimizer~\cite{kingma2014adam} is adopted. The initial learning rate is set to $1 \times 10^{-4}$ and reduced by a factor of 0.3 after every 30 epochs. We train each model for 100 epochs. The shared feature extraction backbone is a ResNet-18 pre-trained on ImageNet. The input size is set to $1024\times1024$. When trained with DocAlign12K, flows from all hierarchical levels and refinement iterations are supervised by the ground-truth flow with $L1$ loss.

\subsection{Comparison with state-of-the-art}

We make comparisons with DGC-Net~\cite{melekhov2019dgc}, Glu-Net~\cite{truong2020glu}, and Glu-GOCor~\cite{truong2020gocor}. All these methods are trained with DocAlign12K. To ensure a fair comparison, we apply our non-rigid pre-alignment method to these methods since DocAlign12K does not consider the background margin. Note that the input spatial resolution for DGC-Net is set to $240\times240$, while that for Glu-Net and Glu-GOCor is set to $1024\times1024$.

We first evaluate DocAligner and the above-mentioned methods on the testing set of DocAlign12K. Similar to dense correspondence in natural images\cite{truong2020glu,truong2020gocor,melekhov2019dgc}, we use average endpoint error (AEPE) and percentage of correct keypoints (PCK)~\cite{yang2012articulated} as metrics. AEPE measures the averaged Euclidean distance between predicted and ground-truth flow fields over all pixels. PCK is defined as the percentage of the correctly estimated points that are within a certain Euclidean distance threshold (in pixels). Results in Table~\ref{tab:epe} demonstrate that DocAligner outperforms previous methods in all metrics by a considerable margin. Particularly, DocAligner obtains a relative improvement of 23.4\% over the second-best method in PCK-1px, indicating DocAligner's ability to achieve precise correlation.

\begin{table}[h]\scriptsize
\caption{Comparisons on DocAlign12K's testing set.}
\label{tab:epe}
\centering
\begin{adjustbox}{width=0.45\textwidth}
\begin{tabular}{cccc}
\hline
           & AEPE$\downarrow$ & PCK-1px (\%)$\uparrow$ & PCK-5px (\%)$\uparrow$ \\ \hline
DGC-Net~\cite{melekhov2019dgc}        & 47.39 & 6.98 & 15.67 \\
Glu-Net\cite{truong2020glu}        & 1.82 & 51.04 & 93.74 \\
Glu-GOCor\cite{truong2020gocor}  & 1.54 & 62.15 & 94.49 \\ \hline
DocAligner & \textbf{1.09} & \textbf{76.63} & \textbf{96.36} \\ \hline
\end{tabular}
\end{adjustbox}
\end{table}

\begin{figure*}[t]
    \centering
    \includegraphics[width=6in]{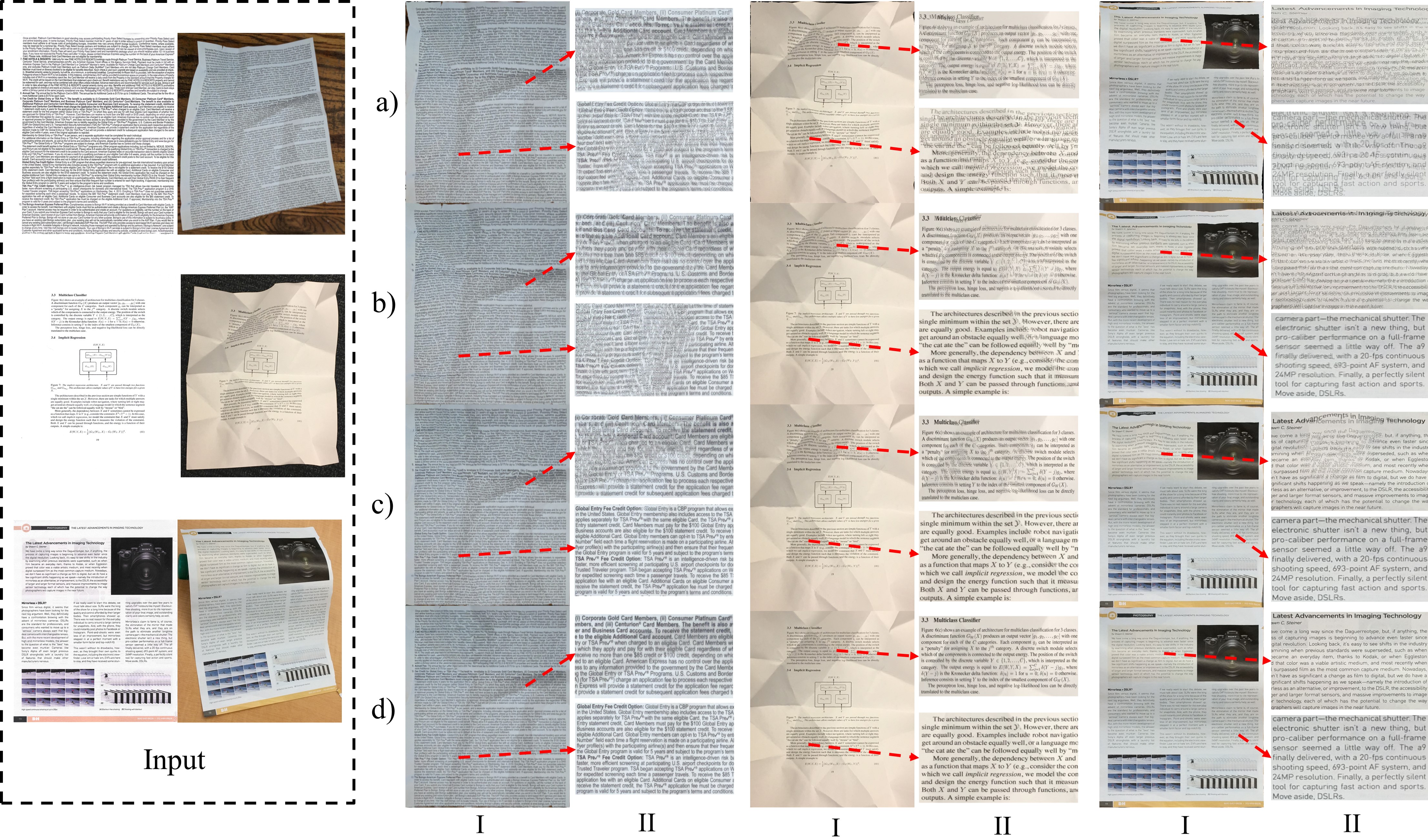}
    \caption{Input pairs are shown on the left. Warped results are shown in I. In II, we overlap targets on warped results and show local details. Top to bottom show results from a) DGC-Net~\cite{melekhov2019dgc}, b) Glu-Net~\cite{truong2020glu}, c) Glu-GOCor~\cite{truong2020gocor}, and d) DocAligner$^{SSFT10}$.}
    \label{fig:results}
\end{figure*}

We then use DocUNet~\cite{ma2018docunet} benchmark to further validate DocAligner' performance, which consists of clean scanned and geometrically-distorted photographic document images. We use the predicted flow field to warp the pre-aligned photographic image: $\tilde{I}'_s = I'_s(\mathbf{x}+f(\mathbf{x}))$, and then assess the alignment between the target clean scanned image $I_t$ and $\tilde{I}'_s$. The better the alignment, the better the model's performance. Multi-scale structural similarity (MS-SSIM)~\cite{wang2003multiscale} and align distortion (AD)~\cite{ma2022learning} are adopted as evaluation metrics. MS-SSIM is a widely used metric that measure perception-based similarity between two images. AD aims to measure the local distortion between two images via dense SIFT flow, which is improved from local distortion (LD)~\cite{you2017multiview} by excluding the noise in low-textured regions and excluding the effect of subtle global transformations. Quantitative and qualitative results are given in Table~\ref{tab:sota1} and Fig.~\ref{fig:results}, respectively. Although DGC-Net can achieve seemingly feasible global alignment, it fails to warp the character details because of the detail-lacking input. The correlation layers in Glu-GOCor are more robust toward repetitive patterns, so it exhibits improvement when compared to Glu-Net. DocAligner achieves superior performance when compared to the above-mentioned methods. Moreover, the gains are further boosted when our self-supervised method is applied. 

\begin{table}[t]
\renewcommand{\arraystretch}{1.1}
\caption{Comparisons with state-of-the-art geometric correspondence methods on DocUNet dataset. ${SSFT10}$ denotes Self-Supervised Fine-Tuning on the entire test set with 10 epochs before testing.}
% Note that the time for fine-tuning is included in the run-time for DocAligner$^{SSFT10}$.}
\label{tab:sota1}
\centering
\begin{adjustbox}{width=0.48\textwidth}
\begin{tabular}{ccccc}
\hline
Methods & MS-SSIM$\uparrow$ & AD$\downarrow$ & Parameters (M)& Run-time (s)\\ \hline
DGC-Net~\cite{melekhov2019dgc} & 0.6177 & 0.3137 & \textbf{68.47} & \textbf{0.48} \\
Glu-Net~\cite{truong2020glu} & 0.7728 & 0.1186
 & 94.17 & 0.75 \\ 
Glu-GOCor~\cite{truong2020gocor}& 0.7862 & 0.0938 & 94.17 & 0.85 \\ \hline
% Cats$++$ ~\cite{shen2020ransac} &  0.6746& 0.4700 & \\ 
DocAligner &  0.8058 & 0.0486
 & 103.8
 & 0.93 \\ 
 % DocAligner$^{TOS25}$ & 0.8190 & 103.8
 % & 38.32 \\ 
 % DocAligner$^{SSFT10}$ & \textbf{0.8232} & \textbf{0.0445} & 103.8 & 6.38 \\ \hline
 DocAligner$^{SSFT10}$ & \textbf{0.8232} & \textbf{0.0445} & 103.8 & 0.93 \\ \hline
\end{tabular}
\end{adjustbox}
\end{table}

In order to validate the effectiveness of our self-supervised approach, we compare it with RANSAC-Flow~\cite{shen2020ransac}, which is another self-supervised method. Similar to the settings in ~\cite{shen2020ransac}, we do not train models on our DocAlign12K but solely adopt the self-supervised training on real data. We train RANSAC-Flow and DocAligner on the WarpDoc~\cite{xue2022fourier} dataset, which consists of 1020 pairs of photographic and clean document images, and then fine-tune and test them on DocUNet. Results shown in Table~\ref{tab:sota2} demonstrate the superiority of  DocAligner. We observe that the poor performance of RANSAC-Flow partially contributes to its rigid pre-alignment, which can not obtain satisfying pre-alignment results. Besides, our self-supervised method is more easy to be implemened compared to sophisticated hierarchical learning for multi-task optimization in RANSAC-Flow.

\begin{table}[t]
\renewcommand{\arraystretch}{1}
\centering
\caption{Performance on DocUNet dataset when solely trained with self-supervision.}
\begin{adjustbox}{width=0.33\textwidth}
\label{tab:sota2}
\begin{tabular}{ccc}
\hline
Methods & MS-SSIM$\uparrow$ & AD$\downarrow$\\ \hline
RANSAC-Flow~\cite{melekhov2019dgc} &  0.6746
 & 0.4700 \\ 
 DocAligner$^{SSFT10}$ & \textbf{0.7864} & \textbf{0.0918} \\ \hline
\end{tabular}
\end{adjustbox}
\end{table}

\subsection{Applications of DocAligner}
To further validate the feasibility and application value of our approach, we annotate photographic document images for document layout analysis (DLA), illumination correction, and geometric rectification using our DocAligner$^{SSFT10}$, and validate the effectiveness of the acquired data. For brevity, we will omit the subscript $SSFT10$ in the following discussion.

\begin{figure}[t]
\includegraphics[width=2.6in,height=1.6in]{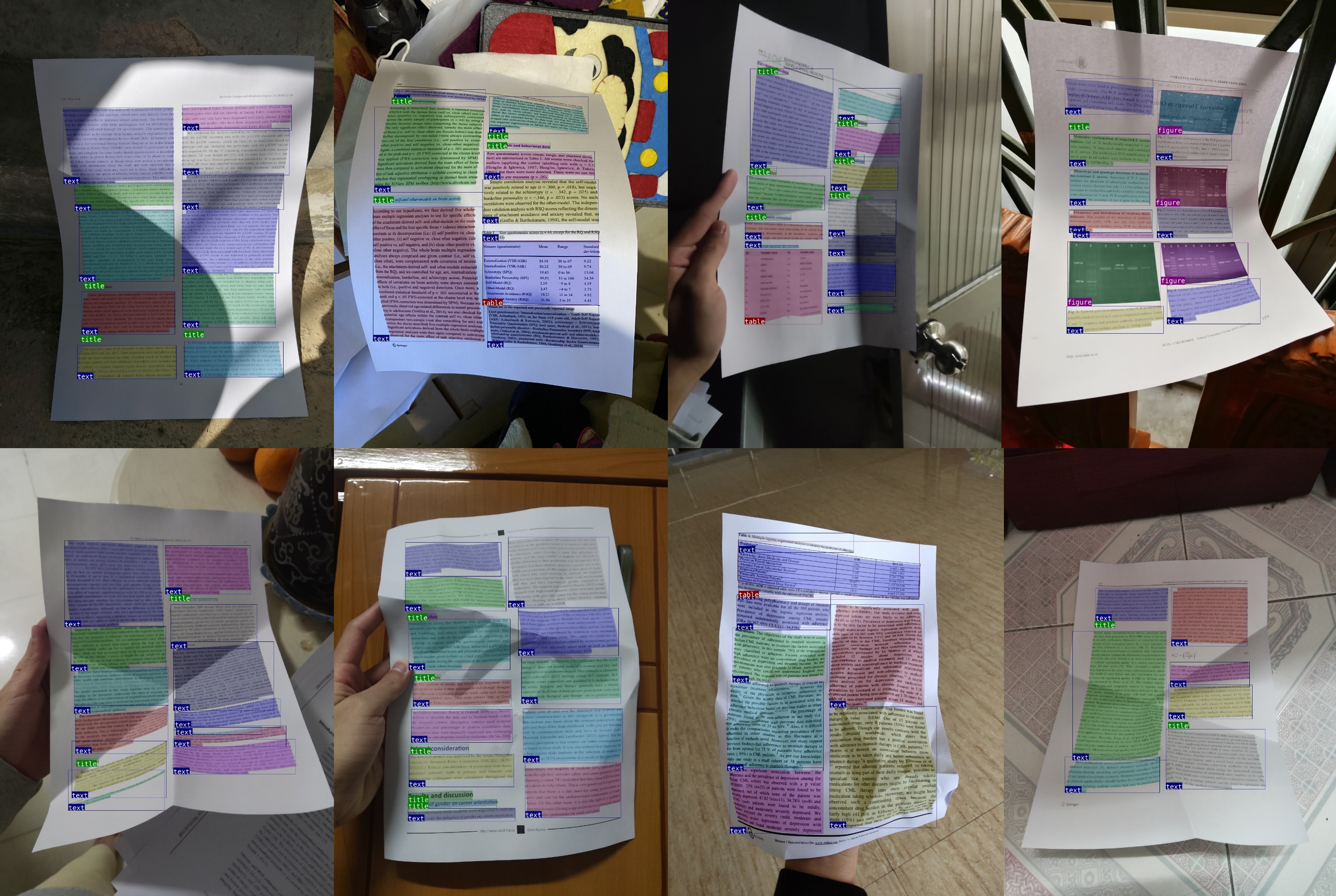}
    \centering
    \caption{Some samples from our acquired DLA dataset.}
    \label{fig:dla}
\end{figure}

\begin{table}[t]\footnotesize
\centering
\renewcommand{\arraystretch}{1.1}
\caption{Performance on photographic testing set when trained with different data. Clean images are original from PubLayNet. S and G represent shadow and geometric synthesis, respectively, the same as DocAlign12K.}
\label{tab:dla}
\begin{tabular}{cccc}
\hline
Training data & Type & Num. & mAP(@0.5-0.95) \\ \hline
Clean images & - &2K& 8.0 \\
Clean images + S & Synthetic &2K& 36.9 \\
Clean images + G & Synthetic &2K& 21.5 \\
Clean images + G + S & Synthetic &2K& 49.7 \\ 
Clean images + G + S & Synthetic &20K& 61.9 \\ \hline
Data from DocAligner & Real &2K& \textbf{68.0} \\ \hline
\end{tabular}
\label{tab:layout}
\end{table}

\textbf{Document layout analysis}. 
We use DocAligner to transfer annotations from an existing dataset for document layout analysis (DLA) to photographic images. To accomplish this, we randomly select 2200 samples from PubLayNet~\cite{zhong2019publaynet}, a large-scale DLA dataset of clean document images. We then print these images and capture them in various environments before using DocAligner to correlate photographic and clean pairs. The resulting flow from DocAligner enabled us to transform the coordinates of bounding boxes and masks to their photographic counterparts, allowing us to obtain the annotations we needed. It is worth noting that traditional manual labeling processes typically cost 5-15 minutes per image, but our approach reduces this to approximately 0.15 minutes per image. Some acquired samples are shown in Fig.~\ref{fig:dla}, which demonstrates DocAligner's ability to generate diverse data with high-quality annotations. To validate the effectiveness of the acquired data, we use it to train a Mask R-CNN~\cite{he2017mask}. We randomly select 200 samples, while the remaining samples form the training set. We manually inspect the testing set and make adjustments to any incorrect annotations to ensure labeling accuracy. The results, shown in Table~\ref{tab:dla}, illustrate the significant superiority of our acquired training data compared to synthetic data. Additional visualization results can be found in the supplementary materials. Furthermore, it is possible to label other detection tasks involving bounding box annotations, such as table and text line detection, using a similar approach.

\begin{table}[t]\footnotesize
\renewcommand{\arraystretch}{1.3}
\caption{Performance on DocUNet dataset when trained with different dataset. Results in the first line represent images without illumination correction (i.e., input images for model).}
\label{tab:illumination}
\centering
\begin{adjustbox}{width=0.44\textwidth}
\begin{tabular}{ccccc}
\hline
Network                 & Training data & Num. & SSIM$\uparrow$   & PSNR$\uparrow$  \\ \hline
-                       & -                & -    & 0.7065 & 12.90 \\ \hline
\multirow{2}{*}{illNet} & DocProj          & 2450 & 0.7139 & 15.74 \\ \cline{2-5} 
                        & Dataset from DocAligner          & 800  & \textbf{0.7504} & \textbf{16.78} \\ \hline
\end{tabular}
\end{adjustbox}
\end{table}

\textbf{Illumination correction}.
\label{illumination}
We utilize our DocAligner to annotate the WarpDoc~\cite{xue2022fourier} dataset, which consists of clean scanned document images and geometrically-distorted photographic document images. However, we exclude the 'Incomplete' subset since our approach does not currently support this document type, as explained in the supplementary material. Furthermore, we exclude images with excessively large rotation angles. It is worth noting that such problematic images can be avoided during the photography process by informing the collectors in advance.
\begin{figure}[t]
    \includegraphics[width=2.5in]{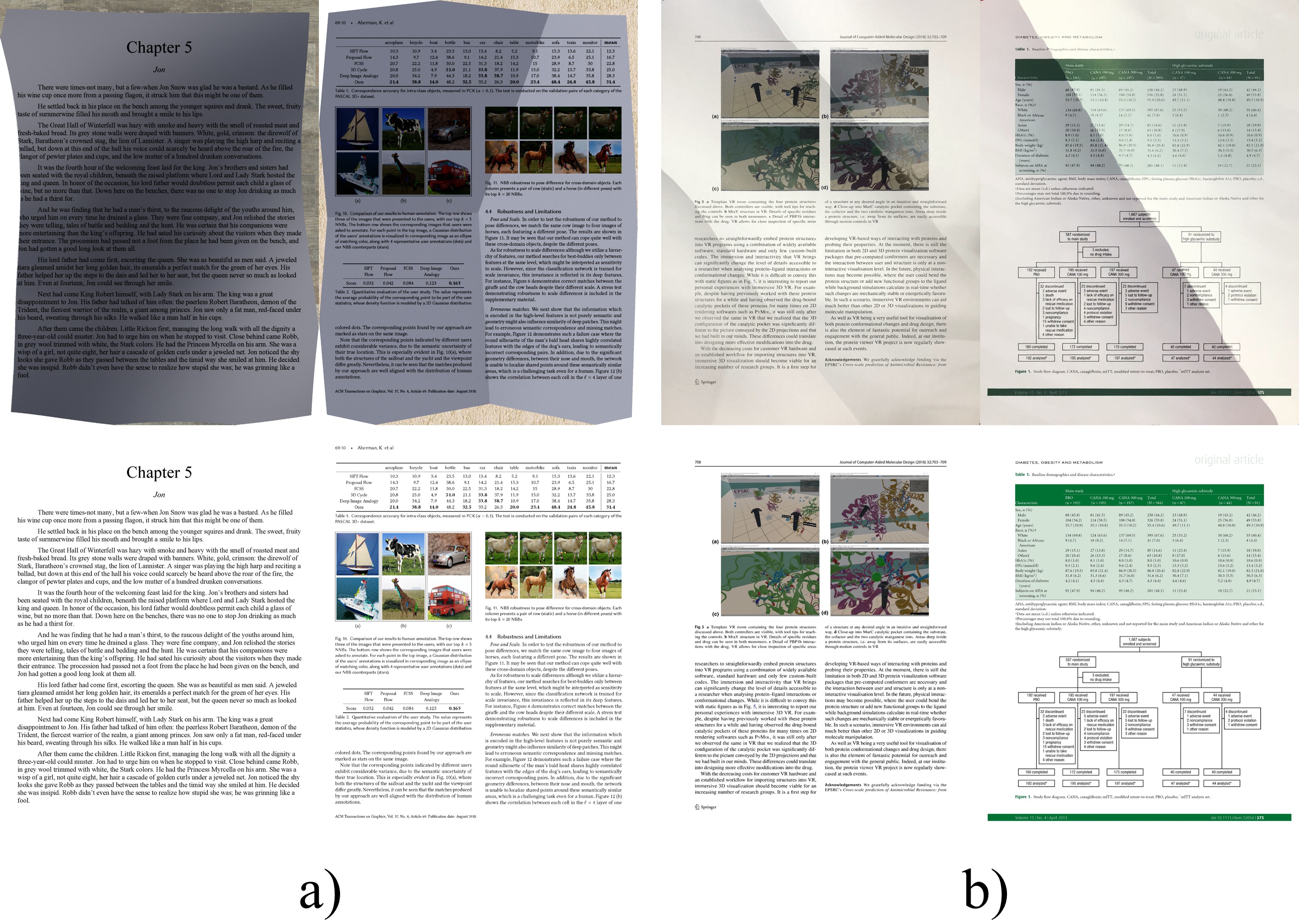}
    \centering
    \caption{a) Examples from synthetic DocProj dataset and b) examples from our acquired real-world dataset.}
    \label{fig:illumination}
\end{figure}
Using the obtained flow field, we warp the photographic source images towards their clean target images to generate paired data for illumination correction. In total, we obtain 800 pairs. Fig.\ref{fig:illumination} presents examples of our acquired data alongside synthetic data from DocProj\cite{li2019document}. It can be observed that the synthetic data lacks diversity and realism. For quantitative comparisons, we train illNet~\cite{li2019document} with our acquired data and DocProj data respectively. The resulting model corrects the illumination of geometrically-rectified DocUNet images from our DocAligner$^{SSFT10}$ (i.e., results depicted in the last row of Table~\ref{tab:sota1}).  Table~\ref{tab:illumination} presents the SSIM and PSNR values between the results obtained by illNet and the corresponding scanned clean images. The results indicate that while model trained with synthetic data yield improvement, it still lag behind model trained with real data, even if the real dataset is smaller in size. Supplementary materials include qualitative comparisons of models trained with different datasets. 

\begin{table}[t]\scriptsize
\centering
\caption{Performance when trained with different data. R and S denote real and synthetic, respectively.}
\label{tab:dewarping}
\renewcommand{\arraystretch}{1.1}
\setlength{\tabcolsep}{0.9mm}{
\begin{tabular}{cccccc}
\hline
Model     & Traning data         & Type           & Num.      & MS-SSIM$\uparrow$ & AD$\downarrow$ \\ \hline
DocUNet~\cite{ma2018docunet}    & Ma et al.~\cite{ma2018docunet}            & S      & 100K      & 0.4157            &  0.4957      \\
DocProj~\cite{li2019document}   &  Li et al.~\cite{li2019document}             &    S         & 1K  &    0.2531          &     0.9278     \\
DDCP~\cite{xie2021document}      & Xie et al.~\cite{xie2021document}           & S      & 30K       & 0.4189            &     0.5071   \\
DewarpNet~\cite{das2019dewarpnet} & Doc3D~\cite{das2019dewarpnet}                & S      & 100K      & 0.4057            & 0.5187         \\
% Marior~\cite{zhang2022marior}    & Doc3D~\cite{das2019dewarpnet}                & S      & 100K      & 0.4267            &                \\
DocTr~\cite{feng2021doctr}     & Doc3D~\cite{das2019dewarpnet}                & S      & 100K      & \underline{0.4649}           & 0.4708         \\
PaperEdge~\cite{ma2022learning} & DIW~\cite{ma2022learning}+Doc3D~\cite{das2019dewarpnet}            & R+S & 2.3K+100K & 0.4523     &    \textbf{0.3901}      \\ \hline
Transformer-based    & \makecell[c]{Dataset from \\ DocAligner} & R           & 2.5K      & \textbf{0.4897}   & \underline{0.4226}           \\ \hline
\end{tabular}}
\end{table}

\textbf{Geometric rectification}. We combine image pairs in DIB~\cite{feng2022geometric}, WarpDoc~\cite{xue2022fourier}, DocUNet~\cite{ma2018docunet}, and our collected DLA data together, from which we randomly select 300 pairs as a testing set. The rest are correlated using DocAligner to get flow fields as annotations. We finally get a training set with 2.5K samples. Inspired by the success achieved by DocTr~\cite{feng2021doctr}, we adopt the Transformer as our dewarping network without extra sophisticated designs. We train it using our constructed training set. As demonstrated in Table~\ref{tab:dewarping}, model trained using our acquired data yield promising results. Our model, based on the vanilla Transformer architecture, surpasses previous meticulously-designed approaches that were trained on large-scale synthetic data. This outcome confirms the effectiveness of the real data we acquired.

\begin{figure}[t]
    \includegraphics[width=2.1in]{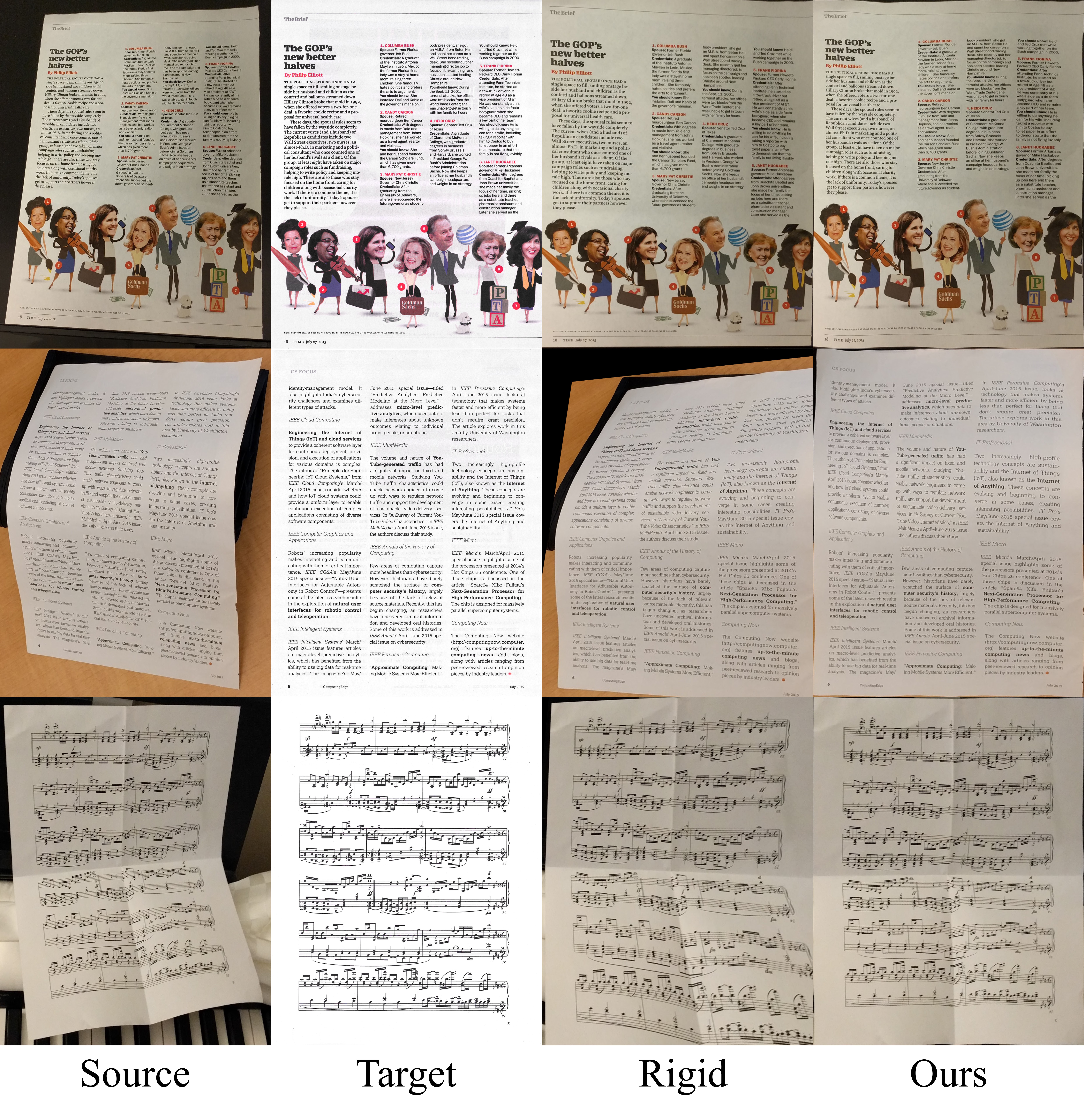}
    \centering
    \caption{Comparisons between results from rigid pre-alignment and our non-rigid pre-alignment.}
    \label{fig:nonrigid}
\end{figure}
\begin{table}[t]\footnotesize
\centering
\caption{Ablation studies on our non-rigid pre-alignment and details recurrent refinement module.}
\label{tab:ablation}
\renewcommand{\arraystretch}{1}
\setlength{\tabcolsep}{1.0mm}{
\begin{tabular}{cccc}
\hline
\makecell[c]{Non-rigid \\pred-alignment} & \makecell[c]{Details \\recurrent refinement} & MS-SSIM$\uparrow$ & AD$\downarrow$  \\ \hline
\XSolidBrush         & \Checkmark                        &   0.7578       &   0.1099  \\
\Checkmark     & \XSolidBrush                      &   0.7910          &    0.0756     \\ 
\Checkmark     & \Checkmark                        & \textbf{0.8058}   & \textbf{0.0486} \\ \hline
\end{tabular}}
\end{table}

\subsection{Ablation studies}
\textbf{Non-rigid pre-alignment}. We replace non-rigid pre-alignment module with homography transformation and evaluate on DocUNet. We use deep features from a pre-trained ResNet-50 to represent two paired images and obtain sparse correspondences based on cosine similarity. Then the RANSAC algorithm~\cite{fischler1981random} is applied to fit a homography. Results in Table~\ref{tab:ablation} indicate that replacing non-rigid pre-alignment leads to significant performance degradation. Some visualized results shown in Fig.~\ref{fig:nonrigid} demonstrate that rigid transformation is only effective in addressing perspective deformation while failing in other cases. In contrast, our non-rigid pre-alignment method proves to be more robust and effective in dealing with various situations.

\textbf{Details recurrent refinement}. To validate the effectiveness of this module, we replace it with another \emph{Local correlation $\mathbf{C_L}$\& flow prediction} block as in Fig.~\ref{fig:architecture} and retrain the model with the same settings. The output size for this variant is one-quarter of the input size, which will be bilinearly up-sampled. Results on the DocUNet dataset in Table~\ref{tab:ablation} demonstrate the improvement introduced by details recurrent refinement.

\section{Conclusions}

We present DocAligner for automating the annotation of photographic document images by establishing dense correspondence between these images and their clean counterparts. DocAligner incorporates several sophisticated techniques, such as non-rigid pre-alignment, hierarchical alignment, details recurrent refinement, a synthesis pipeline, and a self-supervision training approach. Through extensive experiments, we demonstrate the excellent performance of DocAligner in the task of document dense correspondence, generating high-quality annotations. Moreover, DocAligner simplifies the manual annotation process to merely taking pictures, resulting in significant cost reduction associated with manual labeling. When annotating layout analysis data, DocAligner reduces the manual labeling time by a minimum of 30-fold. Experimental results in layout analysis, illumination correction, and geometric rectification emphasize the substantial potential of DocAligner in facilitating various document artificial intelligence tasks in photographic contexts. In our future work, we plan to create additional large-scale real-world datasets for the research community.

{\small
\bibliographystyle{ieee_fullname}
\bibliography{paper}

\begin{thebibliography}{10}\itemsep=-1pt

\bibitem{bi2022srrv}
Hengyue Bi, Canhui Xu, Cao Shi, Guozhu Liu, Yuteng Li, Honghong Zhang, and Jing
  Qu.
\newblock {SRRV}: A novel document object detector based on spatial-related
  relation and vision.
\newblock {\em IEEE TMM}, 2022.

\bibitem{bookstein1989principal}
Fred~L. Bookstein.
\newblock Principal warps: Thin-plate splines and the decomposition of
  deformations.
\newblock {\em IEEE TPAMI}, 11(6):567--585, 1989.

\bibitem{chi2019complicated}
Zewen Chi, Heyan Huang, Heng-Da Xu, Houjin Yu, Wanxuan Yin, and Xian-Ling Mao.
\newblock Complicated table structure recognition.
\newblock {\em arXiv preprint arXiv:1908.04729}, 2019.

\bibitem{cho2022cats++}
Seokju Cho, Sunghwan Hong, and Seungryong Kim.
\newblock Cats++: Boosting cost aggregation with convolutions and transformers.
\newblock {\em IEEE TPAMI}, 2022.

\bibitem{das2019dewarpnet}
Sagnik Das, Ke Ma, Zhixin Shu, Dimitris Samaras, and Roy Shilkrot.
\newblock {DewarpNet}: Single-image document unwarping with stacked {3D} and
  {2D} regression networks.
\newblock In {\em ICCV}, pages 131--140, 2019.

\bibitem{das2020intrinsic}
S Das, H~Ma Sial, R Baldrich, M Vanrell, and D Samaras.
\newblock Intrinsic decomposition of document images in-the-wild.
\newblock In {\em BMVC}, 2020.

\bibitem{dosovitskiy2015flownet}
Alexey Dosovitskiy, Philipp Fischer, Eddy Ilg, Philip Hausser, Caner Hazirbas,
  Vladimir Golkov, Patrick Van Der~Smagt, Daniel Cremers, and Thomas Brox.
\newblock {FlowNet}: Learning optical flow with convolutional networks.
\newblock In {\em ICCV}, pages 2758--2766, 2015.

\bibitem{feng2021doctr}
Hao Feng, Yuechen Wang, Wengang Zhou, Jiajun Deng, and Houqiang Li.
\newblock {DocTr}: Document image transformer for geometric unwarping and
  illumination correction.
\newblock In {\em ACM MM}, pages 273--281, 2021.

\bibitem{feng2022geometric}
Hao Feng, Wengang Zhou, Jiajun Deng, Yuechen Wang, and Houqiang Li.
\newblock Geometric representation learning for document image rectification.
\newblock In {\em ECCV}, pages 475--492, 2022.

\bibitem{fischler1981random}
Martin~A Fischler and Robert~C Bolles.
\newblock Random sample consensus: a paradigm for model fitting with
  applications to image analysis and automated cartography.
\newblock {\em Communications of the ACM}, 24(6):381--395, 1981.

\bibitem{han2017scnet}
Kai Han, Rafael~S Rezende, Bumsub Ham, Kwan-Yee~K Wong, Minsu Cho, Cordelia
  Schmid, and Jean Ponce.
\newblock {SCNet}: Learning semantic correspondence.
\newblock In {\em ICCV}, pages 1831--1840, 2017.

\bibitem{he2017mask}
Kaiming He, Georgia Gkioxari, Piotr Doll{\'a}r, and Ross Girshick.
\newblock Mask {R-CNN}.
\newblock In {\em CVPR}, pages 2961--2969, 2017.

\bibitem{hong2021deep}
Sunghwan Hong and Seungryong Kim.
\newblock Deep matching prior: Test-time optimization for dense correspondence.
\newblock In {\em ICCV}, pages 9907--9917, 2021.

\bibitem{ilg2017flownet}
Eddy Ilg, Nikolaus Mayer, Tonmoy Saikia, Margret Keuper, Alexey Dosovitskiy,
  and Thomas Brox.
\newblock {FlowNet} 2.0: Evolution of optical flow estimation with deep
  networks.
\newblock In {\em CVPR}, pages 2462--2470, 2017.

\bibitem{jeon2018parn}
Sangryul Jeon, Seungryong Kim, Dongbo Min, and Kwanghoon Sohn.
\newblock Parn: Pyramidal affine regression networks for dense semantic
  correspondence.
\newblock In {\em ECCV}, pages 351--366, 2018.

\bibitem{jiang2021cotr}
Wei Jiang, Eduard Trulls, Jan Hosang, Andrea Tagliasacchi, and Kwang~Moo Yi.
\newblock {COTR}: Correspondence transformer for matching across images.
\newblock In {\em ICCV}, pages 6207--6217, 2021.

\bibitem{kaplan2021combining}
Fr{\'e}d{\'e}ric Kaplan, Sofia~Ares Oliveira, Simon Clematide, Maud Ehrmann,
  and Rapha{\"e}l Barman.
\newblock Combining visual and textual features for semantic segmentation of
  historical newspapers.
\newblock {\em Journal of Data Mining \& Digital Humanities}, 2021.

\bibitem{kim2018recurrent}
Seungryong Kim, Stephen Lin, Sangryul Jeon, Dongbo Min, and Kwanghoon Sohn.
\newblock Recurrent transformer networks for semantic correspondence.
\newblock In {\em NeurIPS}, pages 6126--6136, 2018.

\bibitem{kingma2014adam}
Diederik~P Kingma and Jimmy Ba.
\newblock {Adam: A method for stochastic optimization}.
\newblock In {\em ICLR}, 2015.

\bibitem{li2020Tablebank}
Minghao Li, Lei Cui, Shaohan Huang, Furu Wei, Ming Zhou, and Zhoujun Li.
\newblock {TableBank}: Table benchmark for image-based table detection and
  recognition.
\newblock In {\em LREC}, pages 1918--1925, 2020.

\bibitem{li2020docbank}
Minghao Li, Yiheng Xu, Lei Cui, Shaohan Huang, Furu Wei, Zhoujun Li, and Ming
  Zhou.
\newblock {DocBank}: A benchmark dataset for document layout analysis.
\newblock In {\em COLING}, pages 949--960, 2020.

\bibitem{li2019document}
Xiaoyu Li, Bo Zhang, Jing Liao, and Pedro~V Sander.
\newblock Document rectification and illumination correction using a
  patch-based {CNN}.
\newblock {\em ACM TOG}, 38(6):1--11, 2019.

\bibitem{long2021parsing}
Rujiao Long, Wen Wang, Nan Xue, Feiyu Gao, Zhibo Yang, Yongpan Wang, and
  Gui-Song Xia.
\newblock Parsing table structures in the wild.
\newblock In {\em ICCV}, pages 944--952, 2021.

\bibitem{ma2022learning}
Ke Ma, Sagnik Das, Zhixin Shu, and Dimitris Samaras.
\newblock Learning from documents in the wild to improve document unwarping.
\newblock In {\em ACM SIGGRAPH}, pages 1--9, 2022.

\bibitem{ma2018docunet}
Ke Ma, Zhixin Shu, Xue Bai, Jue Wang, and Dimitris Samaras.
\newblock {DocUNet}: Document image unwarping via a stacked {U-Net}.
\newblock In {\em CVPR}, pages 4700--4709, 2018.

\bibitem{melekhov2019dgc}
Iaroslav Melekhov, Aleksei Tiulpin, Torsten Sattler, Marc Pollefeys, Esa Rahtu,
  and Juho Kannala.
\newblock {DGC-Net}: Dense geometric correspondence network.
\newblock In {\em WACV}, 2019.

\bibitem{park2019cord}
Seunghyun Park, Seung Shin, Bado Lee, Junyeop Lee, Jaeheung Surh, Minjoon Seo,
  and Hwalsuk Lee.
\newblock {CORD}: a consolidated receipt dataset for post-ocr parsing.
\newblock In {\em Workshop on Document Intelligence at NeurIPS}, 2019.

\bibitem{pytorch}
Adam Paszke, Sam Gross, Soumith Chintala, Gregory Chanan, Edward Yang, Zachary
  DeVito, Zeming Lin, Alban Desmaison, Luca Antiga, and Adam Lerer.
\newblock Automatic differentiation in {PyTorch}.
\newblock In {\em NeurIPS Autodiff Workshop}, 2017.

\bibitem{pfitzmann2022doclaynet}
Birgit Pfitzmann, Christoph Auer, Michele Dolfi, Ahmed~S Nassar, and Peter~WJ
  Staar.
\newblock {DocLayNet}: A large human-annotated dataset for document-layout
  analysis.
\newblock In {\em ACM SIGKDD}, page 3743–3751, 2022.

\bibitem{rocco2017convolutional}
Ignacio Rocco, Relja Arandjelovic, and Josef Sivic.
\newblock Convolutional neural network architecture for geometric matching.
\newblock In {\em CVPR}, pages 6148--6157, 2017.

\bibitem{shen2020ransac}
Xi Shen, Fran{\c{c}}ois Darmon, Alexei~A Efros, and Mathieu Aubry.
\newblock {RANSAC-Flow}: generic two-stage image alignment.
\newblock In {\em ECCV}, pages 618--637, 2020.

\bibitem{shi2015convolutional}
Xingjian Shi, Zhourong Chen, Hao Wang, Dit-Yan Yeung, Wai-Kin Wong, and
  Wang-chun Woo.
\newblock Convolutional {LSTM} network: A machine learning approach for
  precipitation nowcasting.
\newblock In {\em NeurIPS}, volume~28, 2015.

\bibitem{sun2018pwc}
Deqing Sun, Xiaodong Yang, Ming-Yu Liu, and Jan Kautz.
\newblock {PWC-Net}: {CNNs} for optical flow using pyramid, warping, and cost
  volume.
\newblock In {\em CVPR}, pages 8934--8943, 2018.

\bibitem{teed2020raft}
Zachary Teed and Jia Deng.
\newblock {RAFT}: Recurrent all-pairs field transforms for optical flow.
\newblock In {\em ECCV}, pages 402--419, 2020.

\bibitem{truong2020gocor}
Prune Truong, Martin Danelljan, Luc~V Gool, and Radu Timofte.
\newblock {GOCor}: Bringing globally optimized correspondence volumes into your
  neural network.
\newblock In {\em NeurIPS}, pages 14278--14290, 2020.

\bibitem{truong2020glu}
Prune Truong, Martin Danelljan, and Radu Timofte.
\newblock {GLU-Net}: Global-local universal network for dense flow and
  correspondences.
\newblock In {\em CVPR}, pages 6258--6268, 2020.

\bibitem{vu2021mc}
Xuan-Son Vu, Quang-Anh Bui, Nhu-Van Nguyen, Thi Tuyet~Hai Nguyen, and Thanh Vu.
\newblock {MC-OCR} challenge: Mobile-captured image document recognition for
  vietnamese receipts.
\newblock In {\em RIVF}, pages 1--6, 2021.

\bibitem{wang2021towards}
Jiapeng Wang, Chongyu Liu, Lianwen Jin, Guozhi Tang, Jiaxin Zhang, Shuaitao
  Zhang, Qianying Wang, Yaqiang Wu, and Mingxiang Cai.
\newblock Towards robust visual information extraction in real world: new
  dataset and novel solution.
\newblock In {\em AAAI}, volume~35, pages 2738--2745, 2021.

\bibitem{wang2003multiscale}
Zhou Wang, Eero~P Simoncelli, and Alan~C Bovik.
\newblock Multiscale structural similarity for image quality assessment.
\newblock In {\em ACSSC}, volume~2, pages 1398--1402, 2003.

\bibitem{xie2021document}
Guo-Wang Xie, Fei Yin, Xu-Yao Zhang, and Cheng-Lin Liu.
\newblock Document dewarping with control points.
\newblock In {\em ICDAR}, pages 466--480, 2021.

\bibitem{xue2022fourier}
Chuhui Xue, Zichen Tian, Fangneng Zhan, Shijian Lu, and Song Bai.
\newblock Fourier document restoration for robust document dewarping and
  recognition.
\newblock In {\em CVPR}, pages 4573--4582, 2022.

\bibitem{yang2017learning}
Xiao Yang, Ersin Yumer, Paul Asente, Mike Kraley, Daniel Kifer, and C
  Lee~Giles.
\newblock Learning to extract semantic structure from documents using
  multimodal fully convolutional neural networks.
\newblock In {\em CVPR}, pages 5315--5324, 2017.

\bibitem{yang2012articulated}
Yi Yang and Deva Ramanan.
\newblock Articulated human detection with flexible mixtures of parts.
\newblock {\em IEEE TPAMI}, 35(12):2878--2890, 2012.

\bibitem{you2017multiview}
Shaodi You, Yasuyuki Matsushita, Sudipta Sinha, Yusuke Bou, and Katsushi
  Ikeuchi.
\newblock Multiview rectification of folded documents.
\newblock {\em IEEE TPAMI}, 40(2):505--511, 2017.

\bibitem{zhang2022marior}
Jiaxin Zhang, Canjie Luo, Lianwen Jin, Fengjun Guo, and Kai Ding.
\newblock Marior: Margin removal and iterative content rectification for
  document dewarping in the wild.
\newblock In {\em ACM MM}, pages 2805--2815, 2022.

\bibitem{zhang2021vsr}
Peng Zhang, Can Li, Liang Qiao, Zhanzhan Cheng, Shiliang Pu, Yi Niu, and Fei
  Wu.
\newblock {VSR}: a unified framework for document layout analysis combining
  vision, semantics and relations.
\newblock In {\em ICDAR}, pages 115--130, 2021.

\bibitem{zhong2019publaynet}
Xu Zhong, Jianbin Tang, and Antonio~Jimeno Yepes.
\newblock {PubLayNet}: largest dataset ever for document layout analysis.
\newblock In {\em ICDAR}, pages 1015--1022, 2019.

\end{thebibliography}
}

\clearpage
\noindent\textbf{\Huge Appendix}
\appendix
% \title{Supplementary Materials}
\section{Detailed network architecture}
\begin{figure}[h]
    \includegraphics[width=3.2in]{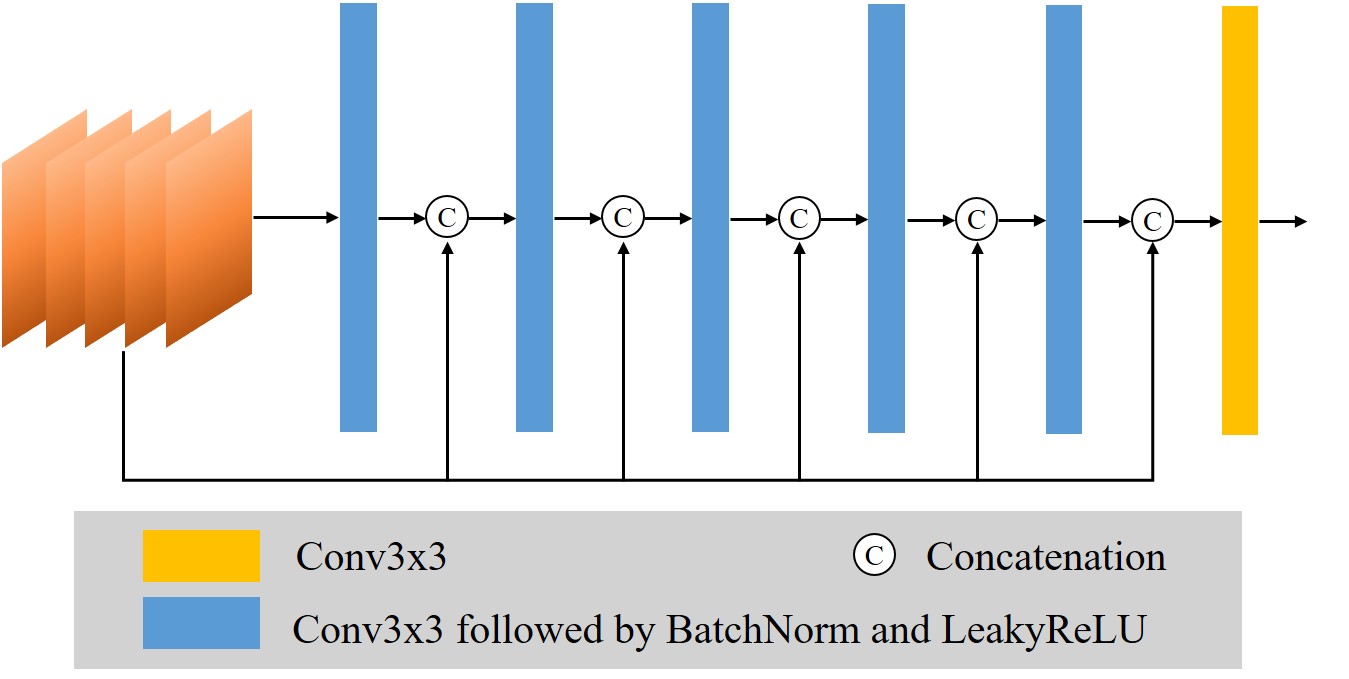}
    \centering
    \caption{Architecture of our flow field decoders.}
    \label{fig:architecture1}
\end{figure}

The architecture of our flow field decoder $\mathbf{deocder}_l()$ described in Equation \color{red}{3} \color{black} is shown in Fig.~\ref{fig:architecture1}, where the output channels for each convolution layer are 128, 128, 96, 64, 32, and 2, respectively. The skip connection will be removed for the lowest resolution level (i.e., $l=1$). 

\begin{figure}[h]
    \includegraphics[width=3.2in]{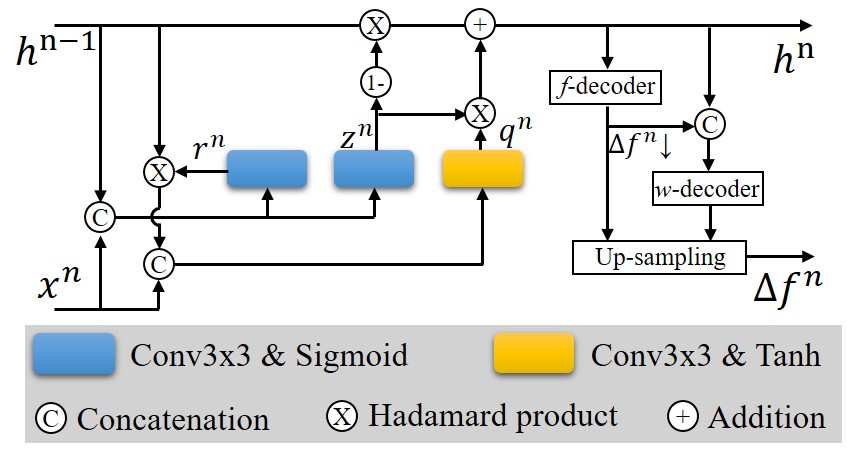}
    \centering
    \caption{Architecture of our ConvGRU unit.}
    \label{fig:gru_details}
\end{figure}

The structure of our ConvGRU unit is depicted in Fig.~\ref{fig:gru_details}. This unit takes an input $x^n$ and a hidden state $h^{n-1}$, and outputs an updated hidden state $h^{n}$ along with a residual flow $\Delta f^n$. The $f$-decoder and $w$-decoder in ConvGRU unit are 2-layer convolutional neural networks with a ReLU activation function. The outputs of $f$-decoder and $w$-decoder are respectively $\Delta  f^n \downarrow \in \mathbb{R}^{\frac{H}{4}  \times \frac{W}{4} \times 2}$ and a weight map $\in \mathbb{R}^{\frac{H}{4}  \times \frac{W}{4} \times 144}$. To obtain the up-sampled residual flow $\Delta f^n$, we first reshape the weight map to a size of $\frac{H}{4} \times \frac{W}{4} \times 16 \times 3 \times 3$, corresponding to 16 weight matrices of size $3\times3$ for each pixel in $\Delta f^n \downarrow$. Then, we calculate each up-sampled pixel value by performing a weighted sum over the $3\times3$ neighborhood in $\Delta f^n \downarrow$ using the $3\times3$ weight matrices from the weight map. Finally, we obtain the desired up-sampled residual flow $\Delta f^n$.

\section{Optimization during self-supervision learning}
As described in the main text, we estimate the solution of the energy function (Equation\color{red}{10} \color{black}) using a gradient descent algorithm, rather than solving it directly. Specifically, we utilize Adam as our gradient descent optimizer with a fixed learning rate of $1\times10^{-4}$. For a more stable optimization, we perform augmentations by warping $I'_{s}$ in each pre-aligned pair $(I'_{s},I_{t})$ three times using different randomly-selected flow fields from DocAligne12K. The resulting augmented pairs, along with the original $(I'_{s},I_{t})$, form a mini-batch of size 4 for optimization during each iteration.

\section{Details on synthesising DocAlign12K}

Our PDF files sourced from magazines, textbooks, scientific articles, and handwritten notes, among others, are predominantly in Chinese and English. We convert these PDF files into images and treat them as target clean document images.

\begin{figure}[h]
    \includegraphics[scale=0.10]{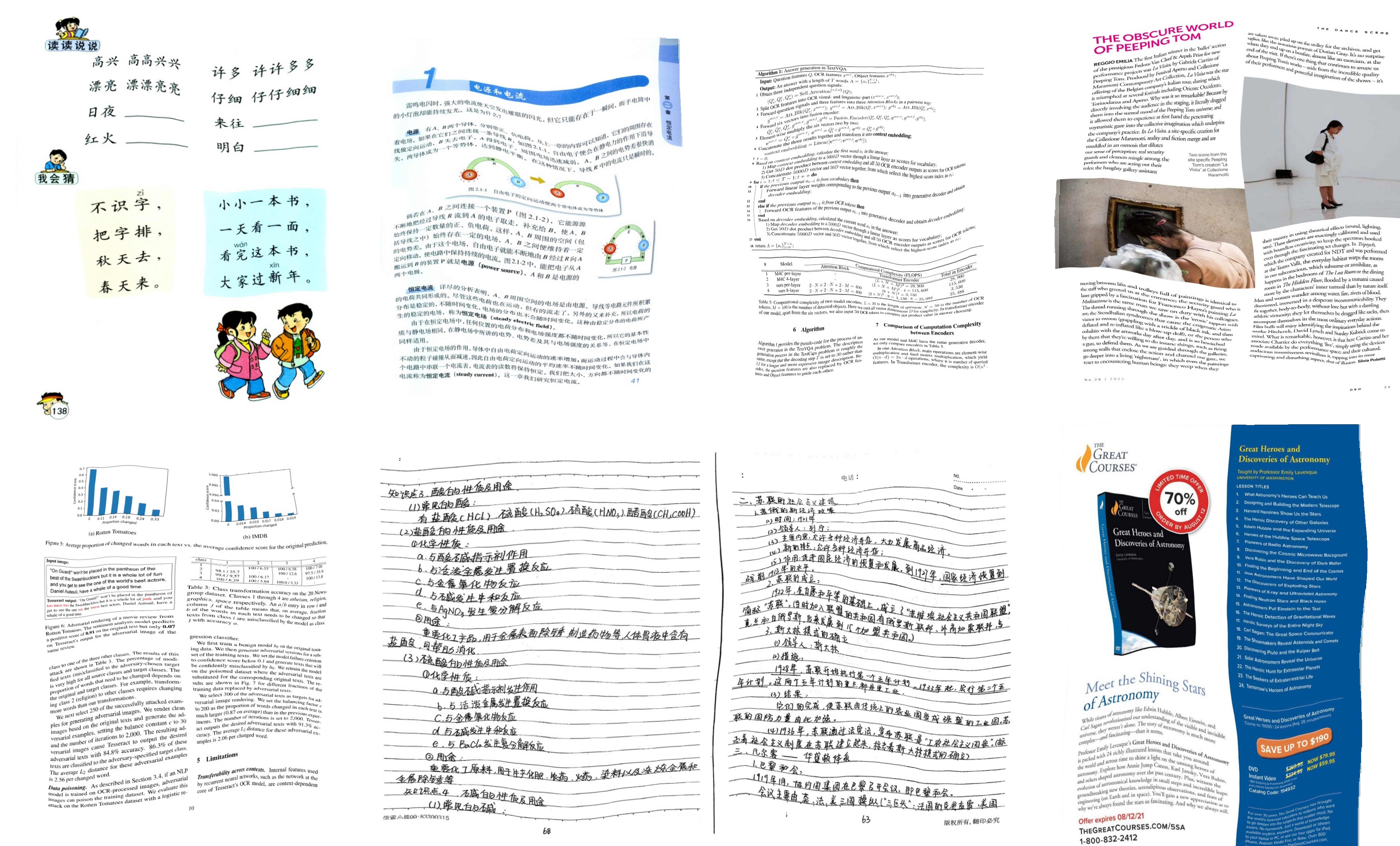}
    \centering
    \caption{Geometrically-distorted document images resulting from our randomly-generated flow fields.}
    \label{fig:synth1}
\end{figure}

We initialize a flow field with a size of $1024\times 1024\times 2$, and then assign a 2-dimensional vector $(dx_i,dy_i)$ for each pixel $i$. Values of $dx_i$ and $dy_i$ are randomly selected from $[-4096, 4096]$ and indicate the direction and distance the pixel is to be shifted. We ensure the flow field is continuous and smooth by applying double mean filtering with a kernel size of 91 each. Although the flow field now contains local perturbations, it still lacks global deformations. Therefore, we add global distortion by considering translation and scaling. Random translation is achieved by respectively adding a random value from $[-50,50]$ to x and y channel in the flow field. For random scaling, we create a base coordinate map $base\_coordicate_{ij} = (i,j) \in \mathbb{R}^{1024 \times 1024 \times 2}$ and a scaling map $scaling_{ij} = (scaling_x,scaling_y) \in \mathbb{R}^{1024 \times 1024 \times 2}$. The values for $scaling\_x$ and $scaling\_y$ are chosen randomly from $[-0.05, 0.2]$. We introduce the scaling deformation to the flow field with the following equation:
\begin{equation}\small
  f = f + (base\_coordicate-512)\otimes scaling,
\end{equation}
where $\otimes$ denotes the Hadamard product and $f$ is our generated flow field. Finally, we use the resulting 2-dimensional flow field to sample geometrically-distorted document images from clean targets. The visualization of our sampled results are shown in Fig. ~\ref{fig:synth1}.

\begin{figure}[t]
    \includegraphics[width=2.6in,height=2.5in]{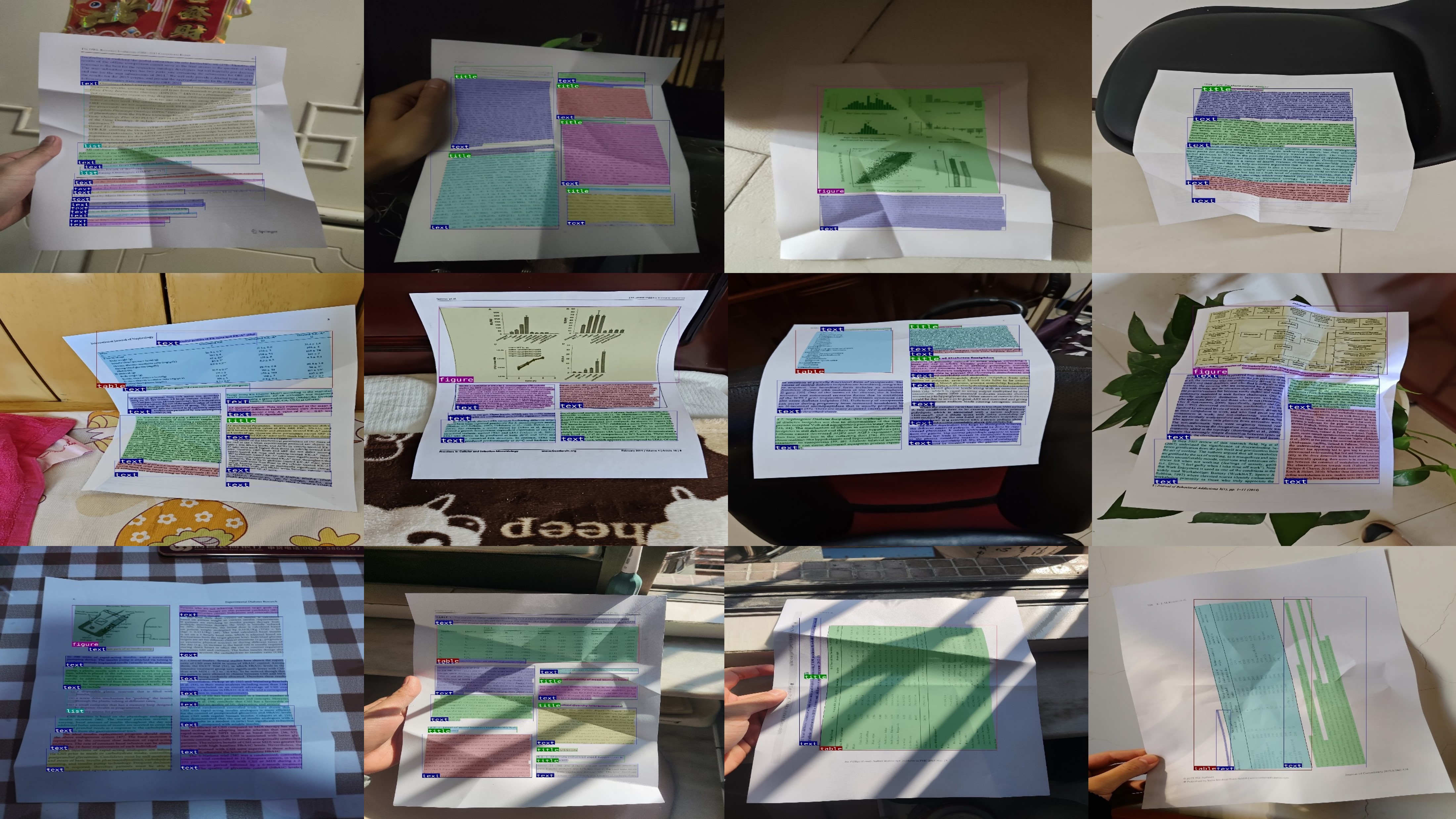}
    \centering
    \caption{More visualized results for our acquired DLA dataset.}
    \label{fig:dla1}
\end{figure}

\section{Additional results for DocAligner's application on layout analysis}

We present additional visualization results for the 2200 samples we acquired for document layout analysis (DLA) in Fig.~\ref{fig:dla1}. These results demonstrate that our DocAligner can produce photographic DLA data that includes various illuminations, geometric deformation, environmental margin, and document content. With high-precision pixel-level alignment, DocAligner can generate accurate and tight-fitting masks and bounding boxes even for geometrically-distorted documents. This is an enormous advantage over manual labeling, which requires a significant amount of time and effort.

\begin{figure}[t]
    \includegraphics[width=2.6in,height=2.5in]{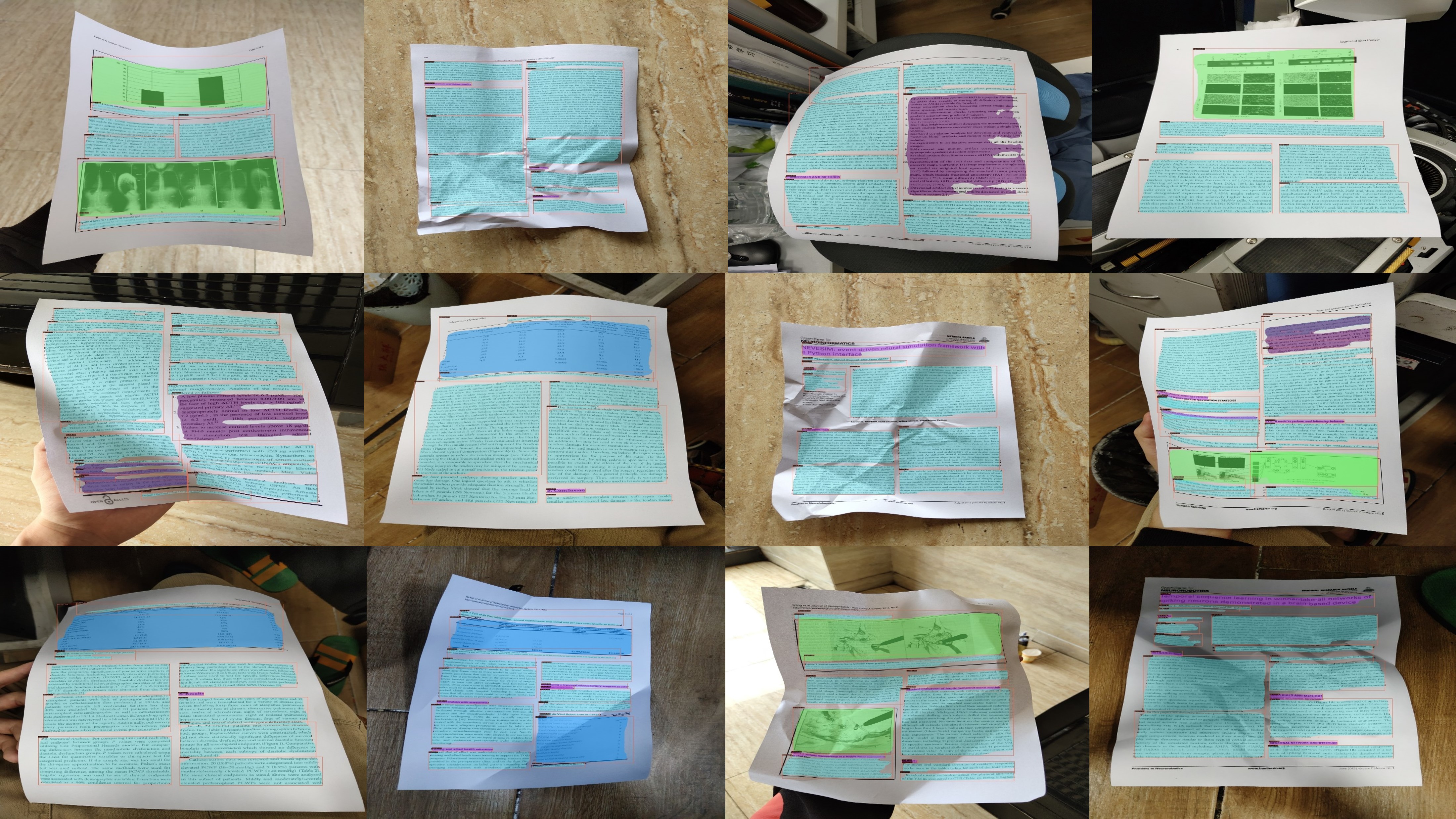}
    \centering
    \caption{More visualized results for our trained Mask R-CNN.}
    \label{fig:dla2}
\end{figure}
\begin{figure}[t]
    \includegraphics[scale=0.11]{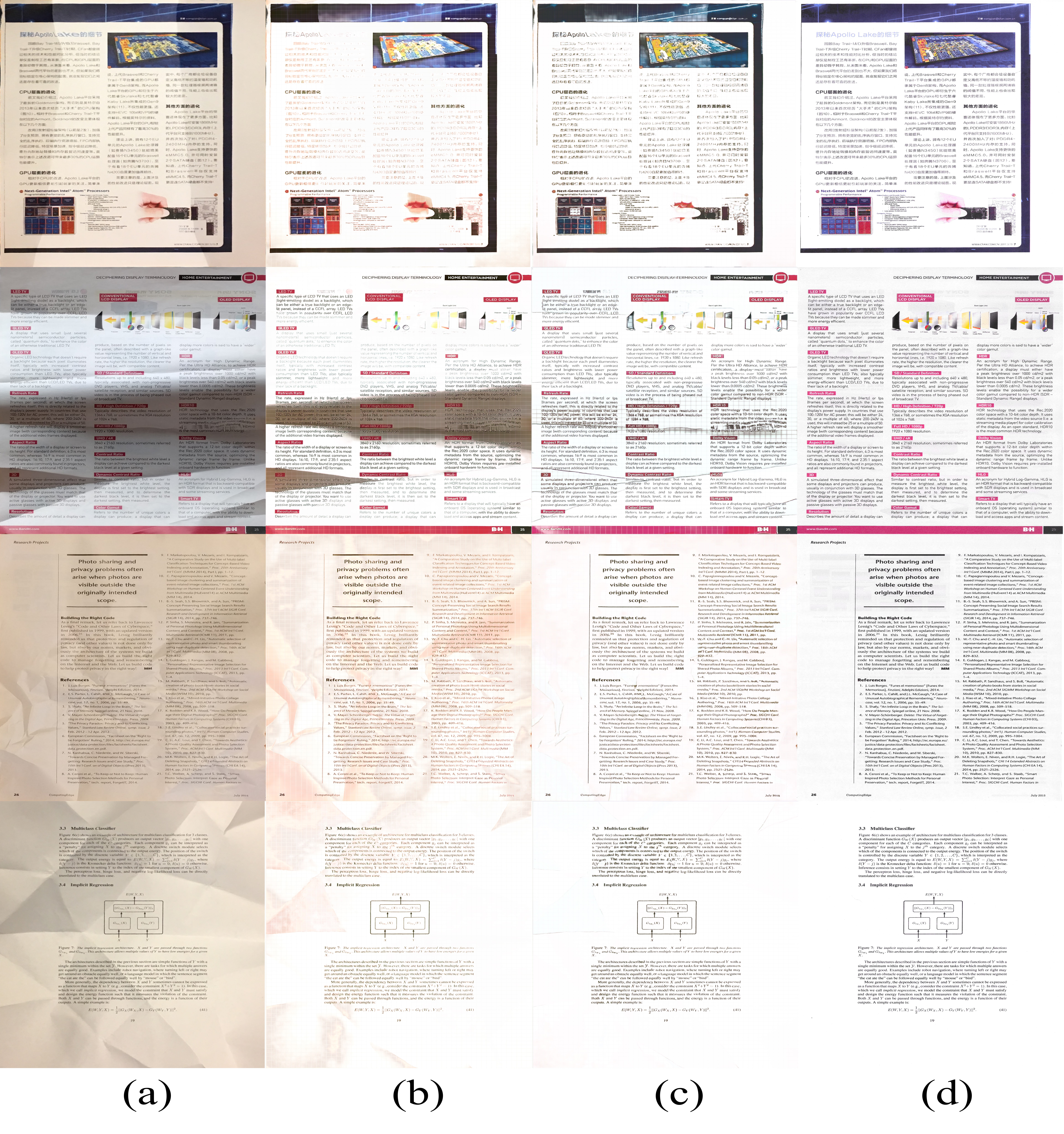}
    \centering
    \caption{From left to right, figures illustrate (a) input images, (b) results from the model trained with the DocProj synthetic dataset, (c) results from the model trained with real-world dataset acquired by DocAligner, and (d) scanned ground-truth.}
    \label{fig:illumination1}
\end{figure}

As described in our main text, we train a Mask R-CNN with ResNet-50 backbone using our collected 2000 training data. Our anchor aspect ratios are set to 0.5, 1, 2, 4, 8, 16, and 24. We train the model on two NVIDIA 2080Ti GPUs with a batch size of 4. We then use the resulting model to inference on 200 test images. Visualized results in Fig.~\ref{fig:dla2} demonstrate that Mask-RCNN trained with DocAligner-acquired data produces suitable performance with such a small scale. This indicates that the acquired data is effective and valuable for further research and development in DLA.

\section{Additional results for DocAligner's application on illumination correction}

As mentioned in Section \color{red}{4.4}\color{black}, we have trained illNet using the synthetic dataset, DocProj, and the real-world dataset, which were acquired by DocAligner. Visualized results are shown in Fig.~\ref{fig:illumination1}. We observe that the model trained using our acquired real-world dataset is more effective in preserving the textual details and reducing the impact of shadows and creases, even though the size of the real-world dataset is relatively small.

\section{Limitations}
\begin{figure}[t]
    \includegraphics[scale=0.35]{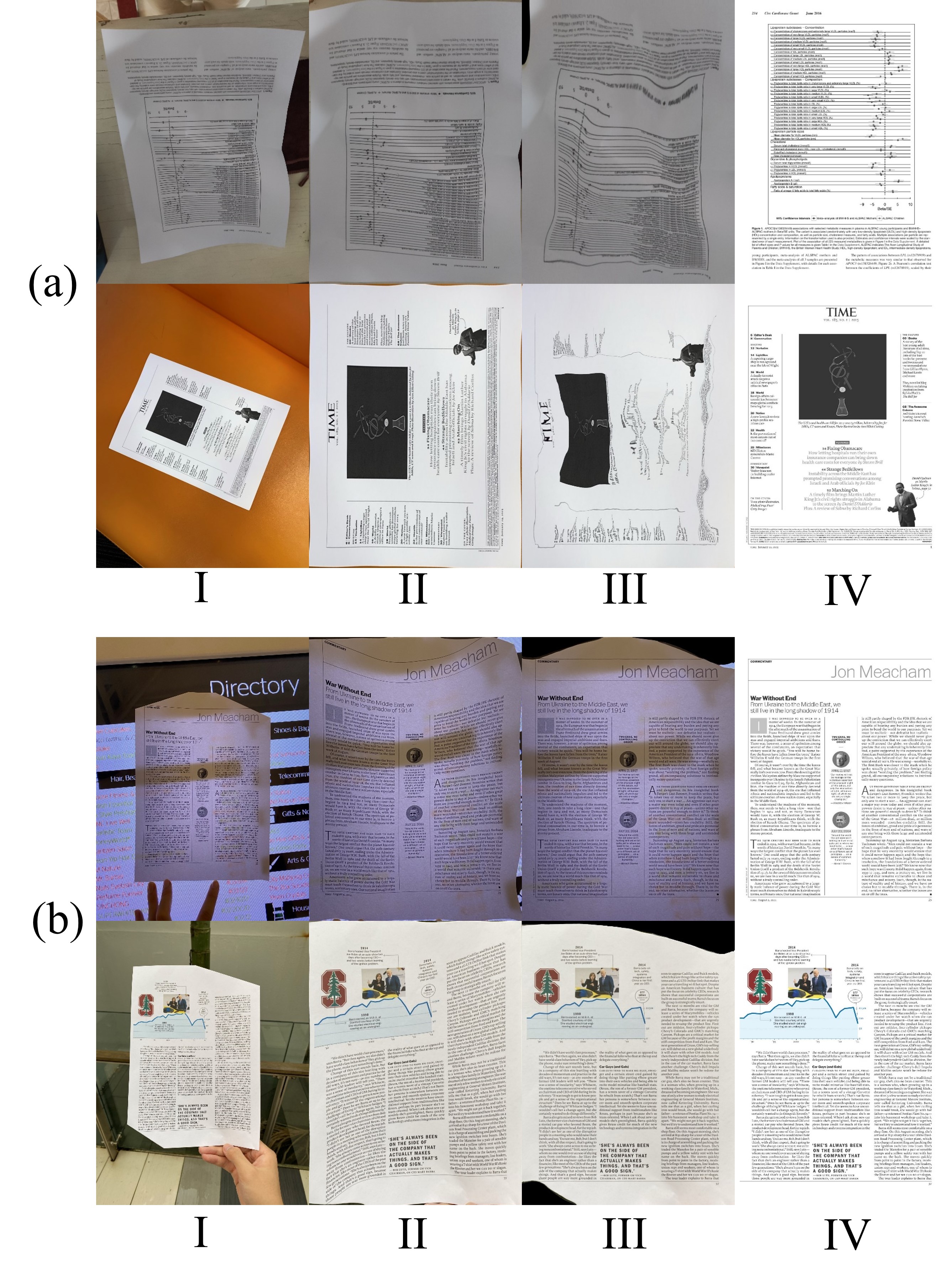}
    \centering
    \caption{Several failure cases encountered during the application of DocAligner. Specifically, (a) illustrates the issue of orientation misalignment, while (b) depicts incomplete documents. From left to right, the figures show \uppercase\expandafter{\romannumeral1} the photographics, \uppercase\expandafter{\romannumeral2} pre-aligned results , \uppercase\expandafter{\romannumeral3} final aligned results, and \uppercase\expandafter{\romannumeral4} target images.}
    \label{fig:failure}
\end{figure}

The current implementation of DocAligner has some limitations, which need to be considered. Firstly, the orientation of photographic images needs to be correct, as our non-rigid pre-alignment module is not orientation aware. Fig.~\ref{fig:failure} (a) illustrates the problematic results produced by DocAligner in pairs containing orientation misalignment. Hence, controlling the camera's attitude to determine the correct orientation is essential. Alternatively, manual adjustment of orientation after capturing is needed.

Secondly, when the source document is incomplete, there may be invalid areas in the aligned results which we fill with zeros. As depicted in Fig.~\ref{fig:failure} (b), such areas make the aligned results unsuitable for document illumination correction tasks.

\end{document}